\documentclass[ejsv2]{imsart}

\RequirePackage[numbers]{natbib}
\RequirePackage[colorlinks,citecolor=blue,urlcolor=blue,allcolors=blue]{hyperref}
\RequirePackage{graphicx}

\startlocaldefs
\theoremstyle{plain}

\newtheorem{theorem}{Theorem}[section]
\newtheorem{lemma}[theorem]{Lemma}
\theoremstyle{definition}

\theoremstyle{remark}

\input{extra_macro.tex}

\def\TITLE{Transfer $Q$-Learning for Finite-Horizon Markov Decision Processes}
\def\SHORTTITLE{Trans$Q$ for Finite-Horizon MDPs}
\endlocaldefs

\begin{document}
\begin{frontmatter}
\title{\TITLE}
\runtitle{\SHORTTITLE}

\begin{aug}
\author[A]{\fnms{Elynn}~\snm{Chen}\ead[label=e1]{elynn.chen@stern.nyu.edu}\orcid{0000-0002-7599-1828}},
\author[B]{\fnms{Sai}~\snm{Li}\ead[label=e2]{saili@ruc.edu.cn}\orcid{0000-0002-6362-3593}}
\and
\author[C]{\fnms{Michael I.}~\snm{Jordan}\ead[label=e3]{jordan@cs.berkeley.edu}\orcid{0000-0001-8935-817X}}
\address[A]{Stern School of Business, New York University, NY, USA\printead[presep={,\\ }]{e1}}

\address[B]{Department of Statistics, Renmin University, Beijing, China\printead[presep={,\\ }]{e2}}
\address[C]{Department of EECS, University of California, Berkeley, CA, USA.\printead[presep={,\\ }]{e3}}
\runauthor{Chen, Li, and Jordan}
\end{aug}

\begin{abstract}
Time-inhomogeneous finite-horizon Markov decision processes (MDP) are frequently employed to model decision-making in dynamic treatment regimes and other statistical reinforcement learning (RL) scenarios. These fields, especially healthcare and business, often face challenges such as high-dimensional state spaces and time-inhomogeneity of the MDP process, compounded by insufficient sample availability which complicates informed decision-making. To overcome these challenges, we investigate knowledge transfer within time-inhomogeneous finite-horizon MDP by leveraging data from both a target RL task and several related source tasks. We have developed transfer learning (TL) algorithms that are adaptable for both batch and online $Q$-learning, integrating valuable insights from offline source studies. The proposed transfer $Q$-learning algorithm contains a novel {\em re-targeting} step that enables {\em cross-stage transfer} along multiple stages in an RL task, besides the usual {\em cross-task transfer} for supervised learning. We establish the first theoretical justifications of TL in RL tasks by showing a faster rate of convergence of the $Q^*$-function estimation in the offline RL transfer, and a lower regret bound in the offline-to-online RL transfer under stage-wise reward similarity and mild design similarity across tasks. Empirical evidence from both synthetic and real datasets is presented to evaluate the proposed algorithm and support our theoretical results.
\end{abstract}



\end{frontmatter}


\section{Introduction}\label{sec:intro}

Transfer learning aims at accelerating the learning process of a target task by leveraging knowledge acquired from different but similar source tasks.
Recently, it has garnered substantial attention in statistical machine learning, due to its empirical successes in domains such as natural language processing, computer vision, game playing, and climate modeling.  On the theoretical front, transfer learning has been investigated within a decision-theoretic framework across a spectrum of supervised learning problems, including classification \citep{cai2021transfer}, high-dimensional linear regression \citep{li2022transfer-jrssb,gu2022robust}, and generalized linear models \citep{li2023estimation}, and there are also applications to problems in unsupervised learning scenarios~\citep{li2022transfer-jasa}.

Our focus is on transfer learning (TL) in the domain of time-inhomogeneous finite-horizon reinforcement learning.  Reinforcement learning (RL) is a sequential decision-making framework that has natural applications in data-driven decision-making settings like robotics, business, healthcare, and education  \citep{levine2020offline,schulte2014q,singla2021reinforcement,kolm2020modern}.
One of the main challenges in RL is that the state space is often of high dimensionality, given that the goal is often that of providing for personalized services or treatments, and given the overall complexity of these domains~\citep{ertefaie2018constructing,luckett2019estimating,ertefaie2021robust,clifton2020q}. Thus there is a significant data scarcity problem in RL.  Transfer learning is a promising framework for addressing data scarcity in RL, with the potential of expediting learning in a target RL task by leveraging related sources tasks from observational, simulated, or offline datasets.  This motivates our interest in integrating of TL perspectives into the realm of RL.

As a first cautionary note, we will demonstrate that a straightforward application of TL algorithms designed for supervised learning (SL) does not yield minimax optimal performance within the RL context. The underlying problem has to do with the distinctive nature of multi-stage optimization and delayed responses in RL, and the cautionary note is equally pertinent in the offline RL setting as it is to the online RL setting.
To address this challenge, we carefully design a novel  framework based on backward inductive $Q$-learning \citep{murphy2005generalization,clifton2020q} and leverages pseudo-responses with re-targeting. It exploits a similarity structure for the reward function, incorporating offline transfer and online $Q$-learning with source offline data while  effectively handling high-dimensional features. We provide a thorough theoretical analysis of this framework for TL in RL, specifically addressing the high dimensionality, and highlighting the distinctive nature of the RL setting compared to TL in SL. 

We describe the contribution of our work in three aspects: new model formulation (for transferable MDPs), novel solution methodology (the re-targeting pseudo response to handle intermediate misalignment), and theoretical guarantees and practical application.
Firstly, under general function approximations, we rigorously define the formulation of TL under finite-horizon MDPs, and unveil a significant distinction between TL in the realms of RL and SL. 
While the concept of {\em ``cross-task transfer''} is present in both SL and RL, allowing knowledge exchange across source and target tasks within the same stage, the notion of {\em ``cross-stage transfer''} is uniquely pertinent to the RL domain. 
This idea taps into future-stage information from source tasks to expedite the learning process in the current stage of the target task. 
While the multi-stage nature of RL tasks is well-acknowledged, its manifestation within TL has remained unexplored. 
In practice, due to the accumulation of differences between source and target tasks over stages, careful consideration is necessary to ensure the transfer's minimax performance. 

This leads us to our second contribution, which involves two novel components integrated into the proposed transfer $Q$-learning algorithm. These components are designed to guarantee minimax optimality and enable both ``cross-task transfer'' and ``cross-stage transfer.''
The first component establishes {\em pseudo-responses} for source tasks through a process called {\em ``re-targeting,''} aligning the future state values of source samples with the corresponding states in the target model. The second component capitalizes on these {\em re-targeted pseudo-responses}, leveraging techniques from supervised transfer learning to estimate the target's optimal $Q^*$-function at each stage. This estimation process utilizes data from both the target task and the pseudo-source data, thus enabling ``cross-task transfer.''

Lastly, our contributions are substantiated through a combination of  rigorous theoretical analysis and empirical evaluation. 
The theory uncovers conditions for effective transfer, offering practical insights. Empirical results affirm the potency of our methods in refining optimal $Q^*$-function estimation for offline RL tasks under reasonable assumptions.  For online RL tasks, our framework demonstrates its value by trimming exploration phases and elevating cumulative rewards, substantiated by empirical findings. This collective evidence underscores the robust practicality of our approach.

\medskip
\noindent
\textbf{Organization.}
The paper is organized as follows:
Section \ref{sec2} presents the model for $Q$-learning and transfer $Q$-learning. 
In Section \ref{sec3}, we introduce our proposed methods for offline and online $Q$-learning with knowledge transfer. 
Theoretical guarantees for our proposals are provided in Section \ref{sec4}. 
Section \ref{sec5} presents an extensive analysis of the numerical performance of our methods in diverse simulation settings and a real-world medical data application. 
Finally, Section \ref{sec6} concludes the paper, summarizing the key findings and highlighting avenues for future research.

\section{Formulation of Transfer between MDP's}\label{sec2}

The mathematical model of an time-inhomogeneous episodic RL task is a finite-horizon MDP defined as a tuple $\calM = \braces{\calS, \calA, \Pr, r, \gamma, T}$, where $\calS$ is the state space, $\calA$ is the action space, $r$ is the reward function, $\gamma\in[0,1]$ is the discount factor, and $T$ is the finite horizon.
At time $t$, for the $i$-th individual, an agent observes the current system state $\bs_{t,i}\in\calS$, chooses a decision $a_{t,i}$ from a finite action set $\calA=\{1, \cdots, M\}$, and transits to the next state $\bs_{t+1,i}$ according to the system transition probability $\Pr\paran{\bs_{t+1,i}|\bs_{t,i}, a_{t,i}}$.
At the same time, she receives an immediate reward $r_{t,i}$, which serves as a partial signal of the goodness of her action $a_{t,i}$.
The reward function $r(\cdot)$ is the expected reward at stage $t$ for the observation $(\bs_{t,i},a_{t,i})$, that is, $r(\bs_{t,i}, a_{t,i}) =\EE[r_{t,i}|\bs_{t,i},a_{t,i}]$.

An agent's decision-making rule is denoted by a policy function $\pi\paran{a_{t,i} | \bs_{t,i}}$ that maps the covariate space $\calS$ to probability mass functions on the action space $\calA$.
For each step $t \in [T]$ and a policy $\pi$, the {\em state-value function} is the expectation of the cumulative discounted reward starting from a state $\bs$ at the $t$-th step:
\begin{equation} \label{eqn:v-func-0}
V^\pi_t(\bs) = \EE^{\pi}\brackets{ \sum_{s=t}^T \gamma^{s-t} r(\bs_{s,i}, a_{s,i}) \bigg| \bs_{t,i} = \bs}.
\end{equation}
The expectation $\EE^{\pi}$ is taken under the trajectory distribution, assuming that the dynamic system follows the given policy $\pi$ afterwards.
Accordingly, the {\em action-value function} or {\em $Q^\pi$-function} of a given policy $\pi$ at step $t$ is the expectation of the accumulated discounted reward starting from a state $\bs$ and taking action $a$:
\begin{equation} \label{eqn:$Q^*$-func-0}
Q^\pi_t(\bs, a) = \EE^{\pi}\brackets{ \sum_{s=t}^T \gamma^{s-t} r(\bs_{s,i}, a_{s,i}) \bigg| \bs_{t,i} = \bs, a_{t,i} = a}.
\end{equation}
For any given action-value function $Q_t^\pi:\calS \times \calA \mapsto \RR$, the {\em greedy policy} $\pi^Q_t$ is defined as,
\begin{equation} \label{eqn:greedy-policy}
\pi^Q_t\paran{a | \bs} =
\begin{cases}
1 &\text{if } a =  \arg\underset{a'\in\calA}{\max}\; Q^\pi_t\paran{\bs, a'},  \\
0 & \text{otherwise}.
\end{cases}
\end{equation}
The overall goal of RL is to learn an optimal policy, $\pi^*_t$, $t\in[T]$, that maximizes the discounted accumulative reward.
To characterize optimality, we define the optimal action-value function $Q^*_t$ as
\begin{equation} \label{eqn:opt-q}
Q^*_t\paran{\bs, a} = \underset{\pi\in\Pi}{\sup}\; Q^\pi_t\paran{\bs, a}, \quad \forall \paran{\bs, a} \in \calS \times \calA.
\end{equation}
The optimal policy $\pi^*_t$ can be derived as any policy that is greedy with respect to $Q^*_t$.
The {\em Bellman optimal equation} holds that
\begin{equation} \label{eqn:bellman-opt}
\EE\brackets{ r_{t,i} + \gamma\; \underset{a'\in\calA}{\max}\; Q^*_{t+1}\paran{\bs_{t+1,i}, a'} - Q^*_t\paran{\bs_{t,i}, a_{t,i}} \bigg| \bs_{t,i}, a_{t,i} } = 0.
\end{equation}
The $Q$-learning algorithms estimate the optimal $Q^*_t$ directly and then obtain the optimal $\pi^*$ as the greedy policy \eqref{eqn:greedy-policy} according to $Q^*_t$. The MDP considered in this paper is time-inhomogeneous since the value functions, specifically $Q^{\pi}_t(\bs, a)$, $Q^*_t(\bs, a)$, and $V^{\pi}_t(\bs)$, depend on the stage index $t$. 
But they are fixed across episodes; we do not assume stochastic drift over episodes.

\subsection{Target and Source MDPs}

Knowledge transfer aims to leverage data from source RL tasks that are {\em similar} to the target task to improve learning of the target RL task.
We consider source data from offline observational data or simulated data.
The target task can be an offline or online RL task. We consider $K$ source tasks from a set $\calK$.  
The target RL task of interest is referred to as the $0$-th task and written with a superscript ``${(0)}$,'' while the source RL tasks are written with a superscript ``${(k)}$,'' for $k\in\calK$. 

Denote the optimal $Q^*$-functions for the target and all source tasks by \\ $Q^{*(k)}_t(\bs,a)$ for $t\in[T]$, $k\in \{0\}\cup\cK$, with the convention that $Q^*_{T+1}(\bs,a)=0$.
The random trajectories for the $k$-th source task are generated from the MDP $\calM^{(k)} = \braces{\calS, \calA, \Pr^{(k)}, r^{(k)}, \gamma, T}$.
Without loss of generality, we assume the horizon length of all tasks are the same and we denote that length as $T$.
For each task $0\le k\le K$, we collect $n_k$ trajectories of length $T$, denoted $\{\bs_{t,i}^{(k)},a_{t,i}^{(k)},r_{t,i}^{(k)}\}_{t\in[T], i\in[n_k]}$. We also assume that the trajectories in different tasks are independent and $n_k$ does not depend on stage $t$, i.e., none of the tasks have missing data.

Single-task RL considers each task $k\in \{0\} \cup \calK$ separately and denote the underlying {\em true response} at step $t$ as
\begin{equation}  \label{eqn-yk}
y^{(k)}_{t,i}
:= r^{(k)}_{t,i}
+ \gamma \cdot \max_{a\in\calA} Q^{*(k)}_{t+1}(\bs_{t+1,i}^{(k)}, \;a).
\end{equation}
According to \eqref{eqn:bellman-opt}, we have
\begin{equation}
\label{bellman0}
\EE\brackets{ y^{(k)}_{t,i} - Q^{*(k)}_{t}\paran{ \bs^{(k)}_{t,i}, a^{(k)}_{t,i} }
\bigg| \bs_{t,i}^{(k)}, a_{t,i}^{(k)} } = 0,
\quad\text{for}\quad k\in \{0\} \cup \calK,
\end{equation}
which provides a moment condition for the estimation of function $Q^{*(k)}_t$.
If $y^{(0)}_{t,i}$ is directly observable, then $Q^{*(0)}_t$ can be estimated via regression.
However, what we observe is a ``partial response'' $r^{(0)}_{t,i}$.
The other component of $y^{(0)}_{t,i}$, as shown in the second term on the RHS of \eqref{eqn-yk}, depends on the unknown $Q^*$-function and future observations. 
As will be discussed in detail in Section \ref{sec3}, we estimate $Q^{*(0)}_t$ in a backward fashion for $t=T,\cdots,1$.

\smallskip
\begin{remark}
In this paper we focus on the case where task similarity is captured primarily through {\em reward differences}. 
Although transition dynamics may also differ across tasks ($\Pr^{(k)} \neq \Pr^{(0)}$), we do not develop full theoretical guarantees for transition transfer here. 
Instead, we sketch several methodological ideas in Section \ref{sec:transition-discussion} and leave a comprehensive treatment to future work.
\end{remark}

\subsection{Similarity Characterization on Reward Functions}

We leverage the similarity of the rewards for the target and source tasks at each stage. Defining RL task difference through the reward function has several advantages. 
Firstly, it is widely applicable to health care, business, and marketing scenarios where individuals may respond slightly differently (in terms of rewards) to the same treatment. 
For instance, in many applications, reward functions represent the revenue generated in a given time period. Discrepancies in revenue across different markets at the same time period are a significant source of divergence in total discounted revenue over a finite time horizon.

Secondly, the similarity assumptions introduced on equation \eqref{eq-delta} can be practically validated, as rewards for all stages are {\em directly observable}.
Lastly, by permitting non-zero difference functions $D^{(k)}(\bs, a)$, our task similarity definition is more generalized compared to that in \cite{lazaric2012transfer} and \cite{mousavi2014context}. In these works, they assume that optimal $Q^*$-functions are identical between different MDPs, which is a more restrictive condition.

Specifically, we define the difference between the reward functions of the target and the  $k$-th source task via
\begin{equation} \label{eq-delta}
\EE\brackets{ r_{t}^{(0)}(\bs_{t,i}, a_{t,i}) - r_{t}^{(k)}(\bs_{t,i}, a_{t,i}) \big| \bs_{t,i} = \bs, a_{t,i} = a}
= D^{(k)}_t(\bs, a),
\end{equation}
for $t\in[T]$ and $k\in[K]$ where $D^{(k)}_t(\bs, a)$ is the the discrepancy function between reward functions.
When this difference, denoted as $D^{(k)}_t(\bs, a)$, is deemed ``small,'' we have the capability to consolidate all source and target data to estimate the ``large'' common component across the data. At the same time, we can utilize only the target data to estimate the ``small'' differences.
We will formally define the magnitude of difference $D^{(k)}_t(\bs, a)$ in Section \ref{sec4}.

The similarity quantification \eqref{eq-delta} also has a potential outcome point of view.
Specifically, for a realized state-action pair $(\bs, a)$, the difference of reward values when switching its participation from the $k$-th task to the target task is $D^{(k)}_t(\bs, a)$.
Because one individual only participates one study in the current setting, the ``switching'' describes unobserved counterfactual facts \citep{kallus2020more}.
Nevertheless, empowered by the potential response framework, we are able to generate counterfactual estimates using samples from $k$-th study and estimated coefficients of the target study.

\section{Transfer $Q$-learning Algorithms for Time-Inhomogeneous MDPs}
\label{sec3}

In Section \ref{sec3-0}, we delineate the novel challenges and phenomena inherent in TL for RL, while also presenting the rationale behind our proposed methods. In Section \ref{sec3-1}, we introduce a $Q$-learning method for offline-to-offline transfer. In Section \ref{sec3-2}, we generalize our proposal to deal with offline-to-online transfer. All discussions are under a general function setting. 

\subsection{New challenge and phenomena in TL for RL}
\label{sec3-0}

We utilize the Bellman optimal equations \eqref{eqn:bellman-opt} and \eqref{bellman0} to estimate the optimal $Q^*$-function for the target task. A fundamental distinction between RL and SL lies in the unavailability of the true response $y^{(0)}_{t,i}$, as defined in \eqref{eqn-yk}.

In the context of single-task $Q$-learning, pseudo-responses $\hat y^{(k)}_{t,i}$ for $k \in {0} \cup [K]$ are typically constructed using the observed $r^{(k)}_{t, i}$ and an estimator of $Q^*_{t+1}(\bs, a)$. 
Thus, an immediate extension of transfer learning (TL) for SL involves augmenting target pseudo-samples $\{\bs^{(0)}_{t,i}, \hat y^{(0)}_{t,i}\}$ with pseudo-samples $\{\bs^{(k)}_{t,i}, \hat y^{(k)}_{t,i}\}$ from the source tasks $k\in\calK$. 
Here, {\em pseudo-response} $\hat y^{(k)}_{t,i}$ is defined as:
\begin{equation}  \label{eqn-pseudo-y}
\hat y^{(k)}_{t,i}
:= r^{(k)}_{t,i} + \gamma\cdot\max_{a\in\calA} Q^{*(k)}_{t+1}(\bs_{t+1,i}^{(k)}, a), \quad\text{for}\quad k\in \{0\} \cup \calK. 
\end{equation}
However, this straightforward construction of pseudo-responses introduces an additional bias due to the mismatch between the source $\max_a\;Q^{*(k)}_{t+1}(\bs,a)$ and the target $\max_a\;Q^{*(0)}_{t+1}(\bs,a)$ on a population level. Specifically, by combining equations \eqref{eqn-yk} and \eqref{eq-delta}, we the following moment equations:
\begin{align}
&\EE \brackets{ y^{(0)}_{t,i}~\big|~\bs_{t,i}=\bs, a_{t,i} =a}
= Q^{*(0)}_{t}(\bs, a),  \label{mm1} \\
&\EE\brackets{ y^{(k)}_{t,i} \cond \bs_{t,i}^{(k)}=\bs, a_{t,i}^{(k)}=a } = Q^{*(0)}_{t}(\bs, a) + D^{(k)}_t(\bs, a) + {\rm Bias}(\bs, a),  \label{mm-k}
\end{align}
where $D^{(k)}_t(\bs, a)$ is typically assumed to be small because of the similarity between tasks, but the additional error 
\begin{align*}
{\rm Bias}(\bs, a) & = \EE\brackets{\gamma\cdot\max_{a\in\calA} Q^{*{(0)}}_{t+1}(\bs_{t+1,i}^{(0)}, a)~\big|~\bs^{(0)}_{t,i}=\bs, a^{(0)}_{t,i} =a} \\
& - \EE\brackets{\gamma\cdot\max_{a\in\calA} Q^{*{(k)}}_{t+1}(\bs_{t+1,i}^{(k)}, a)~\big|~\bs^{(k)}_{t,i}=\bs, a^{(k)}_{t,i}=a}
\end{align*}
resulting from task differences is not inherently minor.

Our innovative approach involves constructing pseudo-responses for the source tasks through a process of {\em ``re-targeting''} them to the target model. 
To grasp the rationale behind this approach, let's consider the population level. 
At any time step $t$, if we possess the true value of $Q^{*(0)}_{t+1}(\cdot,\cdot)$, we can design a {\em ``re-targeted response''} $y^{(tl-k)}_{t,i}$, using source samples, as follows: 
\begin{equation}  \label{eq-mt}
y^{(tl-k)}_{t,i}
:= r^{(k)}_{t,i} + \gamma\cdot\max_{a\in\calA} Q^{*(0)}_{t+1}(\bs_{t+1,i}^{(k)}, a).
\end{equation}
Remarkably, $Q^{*(0)}_{t+1}(\cdot,\cdot)$ is not the true $Q^*$ function of the $k$-th task, but rather that of the target task. 
Notably, $y^{(tl-k)}_{t,i}$ is generated by aligning its future state with the future state of the target model.
By combining equations \eqref{eqn-yk}, \eqref{eq-delta}, and \eqref{eq-mt}, we derive the following moment equation:
\begin{align}
\EE\brackets{ y^{(tl-k)}_{t}~\big|~\bs^{(k)}_{t,i}=\bs, a^{(k)}_{t,i}=a}
= Q^{*(0)}_{t}(\bs, a) + D^{(k)}_t(\bs, a).  \label{mm2}
\end{align}
It is worth-noting that $Q^{*(0)}_{t} + D^{(k)}_t \ne Q^{*(k)}_{t}$ and we do not impose any assumption on $Q^{*(k)}_{t} - Q^{*(0)}_{t}$. 
Comparing \eqref{mm1}, \eqref{mm-k}, and \eqref{mm2}, we can see that, through the re-targeting, the discrepancy between the expected values of $y^{(0)}_{t,i}$ and $y^{(tl-k)}_{t,i}$ solely arises from the inconsistency between their reward functions at stage $t$. 

This rationale underlies the concept of the {\em ``re-targeted pseudo responses''} $\hat y^{(tl-k)}_t$ for the source tasks $k\in[K]$, which is defined by
\begin{equation}  \label{eq-mt-pseudo}
\hat y^{(tl-k)}_{t,i}
:= r^{(k)}_{t,i}
+ \gamma\cdot\max_{a\in\calA} \hat{Q}^{(0)}_{t+1}(\bs_{t+1,i}^{(k)},a),
\end{equation}
where $\hat{Q}^{(0)}_{t+1}$ is an estimate of $Q^{*(0)}_{t+1}$.
For the target task, we use simply the pseudo responses $\hat \by^{(0)}_t$, defined as
\begin{equation}  \label{yk-pseudo}
\hat y^{(0)}_{t,i}
:= r^{(0)}_{t,i} +\gamma\cdot\max_{a\in\calA} \hat{Q}^{(0)}_{t+1}(\bs^{(0)}_{t+1,i},a).
\end{equation}

The premise of transfer learning is the presence of abundant source data and relatively minor differences. Consequently, we can effectively learn the term $Q^{(0)}_{t}(\bs, a) + D^{(k)}_t(\bs, a)$ in \eqref{mm2} by leveraging a substantial amount of source data. Then, due to the small differences, we can also achieve accurate learning of the difference, or bias term, $D^{(k)}_t(\bs, a)$, between \eqref{mm1} and \eqref{mm2}, even with a limited amount of target data. In summary, the transferred estimation for $Q^{(0)}_{t}(\bs, a)$ surpasses single-task estimation due to these factors.

\begin{figure}[t]
\centering
\begin{subfigure}{\linewidth}
\includegraphics[width=\linewidth]{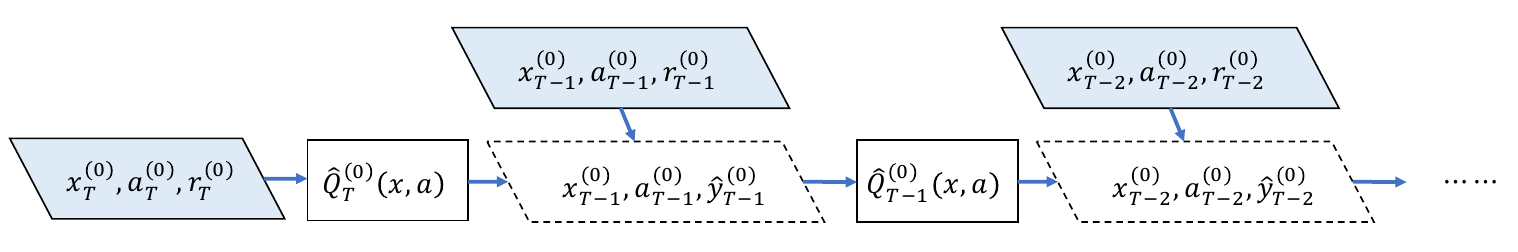}
\caption{Single-task $Q$-learning exclusively employs data from the target task (depicted by blue blocks). The calculation occurs in a reverse manner, and $\hat y_t^{(0)}$ signifies pseudo-responses that are not directly observed.}
\end{subfigure}

\vfill

\begin{subfigure}{\linewidth}
\includegraphics[width=\linewidth]{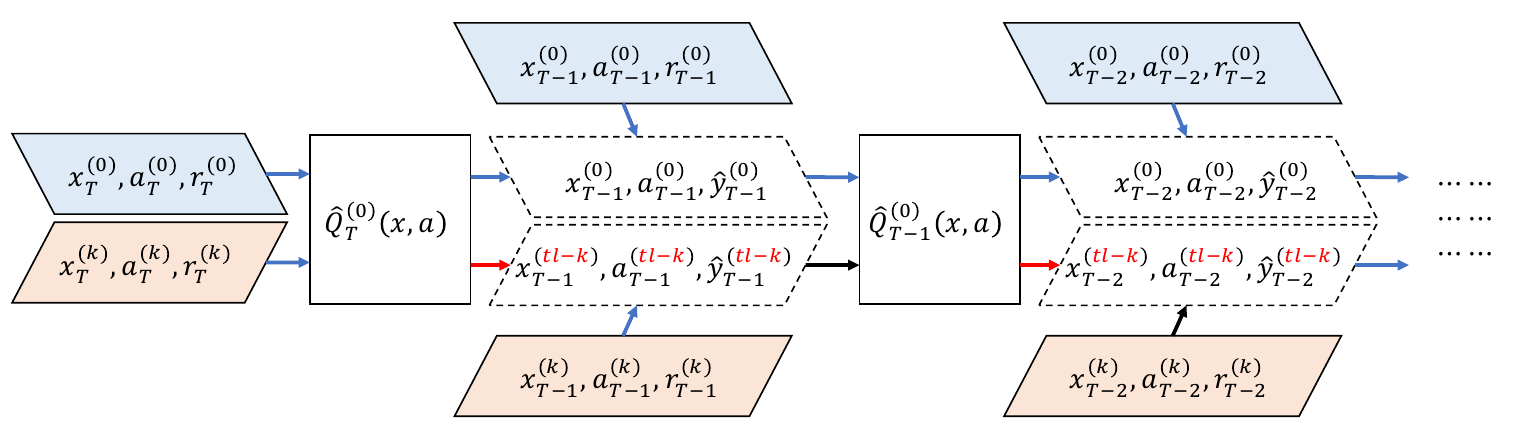}
\caption{Transfer $Q$-learning utilizes data from both the target task (represented by blue blocks) and the $k$-th source task (depicted by orange blocks). $\widehat{Q}_t^{(0)}(x,a)$ stands for the estimated optimal $Q^*$-function at step $t$ for the target task. The re-targeting step (highlighted in red) prevents errors caused by the disparity between $Q_{t}^{*(0)}$ and $Q_{t}^{*(k)}$ from accumulating across stages.}
\end{subfigure}

\caption{\label{fig:transfer}Illustration of single-task $Q$-learning and transfer $Q$-learning.
Naive application of transfer algorithms for regression without re-targeting will incur extra accumulated bias along the red arrows. 
} 
\end{figure}

Figure \ref{fig:transfer} visually outlines the process of our proposed transfer $Q$-learning algorithm in contrast to the conventional single-task $Q$-learning approach. 
It is evident from the illustration that our method employs information aggregation through two key mechanisms.
Firstly, during the estimation of $\hat{Q}^{(0)}_{t}$ at stage $t$, we leverage information from the $t$-th stage of source tasks, represented by $\{\br_{t,i}^{(k)}\}_{k=1}^K$. 
This process is referred to as ``cross-task transfer,'' which is similar to the phenomena of transfer learning for supervised regression.

Secondly, the estimated pseudo response for the target model, denoted as $\hat{\by}_{t,i}^{(0)}$,  is constructed based on the transfer learning estimator $\hat{Q}^{(0)}_{t+1}$. 
This introduces a positive cascading effect across stages: improved accuracy in estimating $\hat{Q}^{(0)}_{s}$, for $t < s \le T$, enhances the estimation of the second term in \eqref{eqn-yk}.
Consequently, this enhancement positively influences the estimation of $\hat{Q}^{(0)}_{t}$ in subsequent steps.
As a direct result, the accuracy of the current-step estimator $\hat{Q}^{(0)}_{t}$ benefits, given that the current response estimation $\hat{\by}_{t,i}^{(0)}$ exhibits lower error compared to its single-task counterpart.

This phenomenon, distinct to RL transfer learning, is termed ``cross-stage transfer,'' and it serves as an additional layer of information exchange that enhances the algorithm's performance.

\subsection{Offline-to-offline transfer}
\label{sec3-1}

Building upon the foundational principle outlined in the previous section, we introduce a transfer $Q$-learning algorithm that follows a backward inductive approach, designed specifically for offline-to-offline transfer. 

The algorithm's flow is summarized in Algorithm \ref{alg-master}. 
Starting from the final stage $T$ where we have full observations of $\{\bs_{T,i}^{(k)},a_{T,i}^{(k)},y_{T,i}^{(k)}\}_{i\in[n_k]}$ for each task, we apply supervised transfer learning algorithms to estimate the target function $\hat{Q}^{(0)}_T$.
Proceeding backwardly to stages $t = T-1, \cdots, 1$, we calculate the pseudo response $y_{t,i}^{(0)}$ in \eqref{yk-pseudo} using $\hat{Q}^{(0)}_{t+1}$, thereby estimating the $Q^*$-function's observations for the target task at stage $t$.
To leverage the source data, we calculate the re-targeted pseudo response $y_{t,i}^{(tl-k)}$ in \eqref{eq-mt-pseudo} with the help of $\hat{Q}^{(0)}_{t+1}$.

The general transfer learning procedure involves two main steps: 
(i) It begins by estimating a pooled estimator from the combined source re-targeted pseudo data $y_{t,i}^{(tl-k)}$ for $k\in[K]$; (ii) It then estimates the bias using the target pseudo data $y_{t,i}^{(0)}$ and estimated pooled $\hat{Q}$. 
The final estimate $\hat{Q}^{(0)}_{t+1}$ is obtained by debiasing the pooled estimator with the estimated bias.

The penalty term $\lam_0\cdot p(D)$ serves to enforce the ``small'' difference assumption, while the penalty term $\lam_{\aux}\cdot p(Q)$ prevents overfitting of the pooled $Q^*$ function.

\begin{algorithm}[tb]
\SetKwInOut{Input}{Input}
\SetKwInOut{Output}{Output}
\Input{Target data $\{\{\bs^{(0)}_{t,i}, a^{(0)}_{t,i},r^{(0)}_{t,i}\}_{t\in [T]}\}_{i\in[n_0]}$, \\
source data $\{\{\{\bs^{(k)}_{t,i},a^{(k)}_{t,i},r^{(k)}_{t,i}\}_{t\in[T]}\}_{i\in[n_k]}\}_{k=1}^K$, and \\
discount factor $\gamma\in[0,1]$.}

Let $\hat{Q}^{(0)}_{T+1}(\cdot)=0$ to deal with pseudo-response construction at last stage $T$. 

\For{$t=T,\dots, 1$ \tcp*{Backward calculation from the last stage.}}{  

Construct re-targeted pseudo-response $\hat y^{(tl-k)}_{t,i}$ \eqref{eq-mt-pseudo} and $\hat y^{(0)}_{t,i}$ \eqref{yk-pseudo}. 

\tcc{Supervised regression transfer block: aggregate and debias.}
Transfer learning:  
\begin{equation*}
\begin{aligned}
\hat{Q}_t&=\argmin_{Q} \left\{\sum_{k=1}^K \sum_{i=1}^{n_k}\ell\paran{\hat{y}_{t,i}^{(tl-k)}-Q(\bs^{(k)}_{t,i}, a^{(k)}_{t,i})}+\lam_{\aux}\cdot p(Q)\right\}.\\
\hat{D}_t&=\argmin_{D} \left\{\sum_{i=1}^{n_0}\ell\paran{\hat{y}_{t,i}^{(0)}-\hat{Q}_t(\bs^{(0)}_{t,i}, a^{(0)}_{t,i}) - D(\bs^{(0)}_{t,i}, a^{(0)}_{t,i})}+\lam_0\cdot p(D)\right\}.
\end{aligned}
\end{equation*}
where $\ell(\cdot)$ and $p(\cdot)$ are any suitable loss and regularization functions chosen for desired scenarios.  

Set $\hat{Q}^{(0)}_t=\hat{Q}_t+\hat{D}_t$.
}

\Output{$\hat{Q}^{(0)}_t$ for all stage $t \in [T]$.}

\caption{Transfer $Q$-learning (General)}
\label{alg-master}
\end{algorithm}

\medskip
\noindent
\textbf{Adaptability of the supervised regression transfer block.}
The component of the supervised regression transfer block is designed to be flexible, allowing for seamless substitution with various transfer learning methodologies from supervised learning, contingent upon their assumptions and objectives \citep{tripuraneni2020provable,bastani2021predicting,gu2022robust,li2023estimation}. An instance of transfer $Q$-learning with linear $Q^*$ and TransLASSO \cite{li2022transfer-jrssb} is detailed in Section \ref{sec4} and Algorithm \ref{alg-linear}. 

\subsection{Offline-to-online transfer}
\label{sec3-2}
Now, we introduce the offline-to-online transfer $Q$-learning algorithm.
In this context, transfer learning can provide an efficient way to leverage the offline data, including demonstrations for the target task, suboptimal policies and demonstrations for the source tasks.
Algorithm \ref{alg-online} outlines the online transfer $Q$-learning framework, grounded in the exploration and then commit (ETC) paradigm.

\begin{algorithm}[t]
\SetKwInOut{Input}{Input}
\SetKwInOut{Output}{Output}
\Input{Offline source data $\{\{\{\bs^{(k)}_{t,i},a^{(k)}_{t,i},r^{(k)}_{t,i}\}_{t\in[T]}\}_{i\in[n_k]}\}_{k=1}^K$.\\
Exploration phase length $n_e$, an exploration policy $\pi_e$.}

{\color{blue} \# Exploration phase}

\For{$i=1,\dots, n_e$}{
\For{$t=1,\dots, T$}{
Take action $a_{t,i}^{(0)}=\pi_e(\cdot|\bs^{(0)}_{t,i})$, get $r^{(0)}_{t,i}$ and $\bs^{(0)}_{t+1,i}$.
}
}
{\color{blue} \# Transfer phase}

Run Algorithm \ref{alg-master} with target data $\{\{\bs^{(0)}_{t,i},a^{(0)}_{t,i}, r^{(0)}_{t,i}\}_{t\in[T]}\}_{i\in[n_e]}$ and offline source data $\{\{\{\bs^{(k)}_{t,i},a^{(k)}_{t,i},r^{(k)}_{t,i}\}_{t\in[T]}\}_{i\in[n_k]}\}_{k=1}^K$. 

Output $\{\hat{Q}^{(0)}_t\}_{t=1}^T$.

{\color{blue} \# Exploitation phase}

\For{$i=n_e+1,\dots, $}{
\For{$t=1,\dots, T$}{
Take greedy action $\widehat{a}_{t,i}=\argmax_{a\in[M]} \hat{Q}^{(0)}_t(\bs^{(0)}_{t,i},a)$ and transit to   $\bs^{(0)}_{t+1,i}$.
}
}
\Output{Online actions $\{\widehat{a}^{(0)}_{t,i}\}_{t\in[T]}$, ~$i=1,\dots$}
\caption{Explore-Transfer-Then-Commit: Offline to Online Transfer}
\label{alg-online}
\end{algorithm}

Algorithm \ref{alg-online} first generates $n_e$ trajectories from the target task in the exploration phase.
Then in the learning phase, Algorithm \ref{alg-master} is called to combine the target data with the offline source data to estimate the target parameters.
Lastly in the exploitation phase, it executes a greedy action to maximize the estimated $Q^*$-function at each stage afterwards.
The longer the exploration phase, the larger the regret since the estimated optimal policy is not employed in the exploration phase.
We will show in theory that $n_e$ can be set much smaller by leveraging offline data under certain conditions.
Specifically, it suffices to take $n_e\geq c\log p$ in Algorithm \ref{alg-online} when having a large amount of offline data. In contrast, without offline data, we show that one needs to take $n_e\geq cs\log p$.
This is because one can learn a near-optimal policy based on the offline data when the offline tasks are sufficiently similar to the target task.

\subsection{Streaming Retargeting for Online Sources}
\label{sec:streaming}

The re-targeting step in Algorithm~1 requires only the observed transition 
$(s_t^{(k)},a_t^{(k)},r_t^{(k)},s_{t+1}^{(k)})$ and the current estimate of the target value 
$\widehat Q^{(0)}_{t+1}$:
\[
\hat{y}^{\,tl-k}_{t,i} \;=\; r^{(k)}_{t,i} + \gamma \max_{a\in\calA} 
\widehat Q^{(0)}_{t+1}(s^{(k)}_{t+1,i},a).
\]
Thus, re-targeting never requires observing actions beyond $t+1$. This observation 
allows us to adapt the method efficiently to streaming settings.

\smallskip
\noindent
\textbf{Episode-buffered approach.}
A straightforward strategy is to buffer full episodes from each source and, once an 
episode terminates, run a backward pass to re-target labels and update estimates. 
While statistically valid, this strategy is computationally expensive, as it requires 
waiting for complete episodes.

\smallskip
\noindent
\textbf{One-step re-targeting.}
In streaming sources, a more efficient approach is to form re-targeted labels online. 
As soon as a transition is completed, we immediately compute the re-targeted label 
using the current $\widehat Q^{(0)}_{t+1}$ and add it to a per-stage buffer. 
This requires only a one-step delay (until $s_{t+1}$ is observed), not the full episode.

\smallskip
\noindent
\textbf{Mini-batch backward updates.}
We maintain stage-wise buffers $\cB_t^{\text{src}}$ and $\cB_t^{\text{tgt}}$ of recent tuples 
and their re-targeted labels. Periodically (e.g., every $H$ new target episodes or when 
$\widehat Q^{(0)}$ changes materially), we perform a short backward sweep over 
$t=T,\ldots,1$: 
(i) re-label only the right-hand sides (since feature matrices $W_t$ remain fixed), and 
(ii) update the pooled and de-bias regressions using warm starts (Algorithm~3 in the 
linear case). This procedure preserves the ``cross-stage'' backward structure while 
avoiding full-episode relabeling, yielding substantial computational savings.

In summary, one-step streaming re-targeting ensures that our framework extends naturally 
to continuous data arrival, while maintaining both statistical validity and computational 
efficiency.

\subsection{Beyond Reward Similarity} \label{sec:transition-discussion}

Our current framework focuses on stage-wise reward similarity with mild design similarity across tasks, and does not provide guarantees under general transition shifts. While this setting is broad and practically verifiable, important extensions remain. In particular, when transition dynamics differ significantly across tasks, re-targeting alone may leave residual bias. 

We briefly outline several promising directions for handling such transition shifts:  
\begin{itemize}
\item \textbf{Importance-weighted re-targeting.} By estimating conditional transition density ratios
$\omega^{(k)}_t(s' \mid s,a) \;=\; \frac{p^{(0)}_t(s' \mid s,a)}{p^{(k)}_t(s' \mid s,a)}$, 
source transitions can be re-weighted so that their conditional distribution aligns with the target’s. This corrects for covariate shift in next-state features and provides unbiased re-targeted samples when transitions differ. 

\item \textbf{Representation alignment.} In linear and feature-based MDPs, one can estimate or learn successor features that capture long-run transition dynamics. Aligning these representations across tasks, through penalization or constrained regression, allows pooled estimation to remain valid even under transition heterogeneity.

\item \textbf{Robust transfer via distributional penalties.} Rather than assuming exact similarity, one can add penalties based on integral probability metrics (IPMs) such as Maximum Mean Discrepancy (MMD) or Wasserstein distance to discourage reliance on sources whose next-state feature distributions diverge from the target \cite{sriperumbudur2009integral,gretton2012kernel}. This yields error bounds with explicit additive terms reflecting the magnitude of transition shift.
\end{itemize}
These extensions call for new estimation tools and theoretical analysis, but they provide a principled roadmap for expanding our framework beyond the reward-similarity regime studied here. 
Recent papers in transfer learning in RL has explored some of those directions \cite{chen2024data,chai2025deep,zhang2025transfer,chai2025transfer,zhou2025prior}. 
A full development of these methods is beyond the scope of this paper and will be pursued in future work.

\section{Theoretical Results}
\label{sec4}

For our theoretical analysis we focus on the \emph{linear function approximation} setting (Algorithm~\ref{alg-linear}), which is widely adopted in applications across health, economics, and business. 
Recall that Algorithm~\ref{alg-master} specifies the general transferred $Q$-learning framework; in this section we analyze its \emph{linear instantiation}, Algorithm~\ref{alg-linear}. 
All formal guarantees, including Theorem~\ref{thm1-tl}, are therefore stated for Algorithm~\ref{alg-linear}. 
While the overall proof strategy extends in spirit to semi-parametric, nonparametric, or other general function approximation classes, such cases require substantially different technical tools. 
We view these extensions as important directions for future research.

The optimal action-value and the reward difference function are parameterized as
\begin{align}
& Q_t^{*(k)}(\bs, a) = Q_t^*(\bs, a;\;\btheta^{(k)}) = \bw(\bs, a)^{\top} \btheta^{(k)}_t,\label{eqn:q-k-linear}\\
& D_t^{(k)}(\bs, a) = D_t(\bs, a;\;\bdelta^{(k)}) = \bw(\bs, a)^{\top} \bdelta^{(k)}_t,  \label{eqn:delta-k-linear}
\end{align}
for $k\in\{0\}\cup[K]$ where $\bw(\bs, a)=[\bphi(\bs)^{\top} \bbone(a=1), \cdots, \bphi(\bs)^{\top} \bbone(a=M)]$, $\phi(\cdot)$ are fixed nonlinear feature functions and can be viewed as the state representation. 
Linear $Q^*$-functions on features have been widely used in literature of dynamic treatment regime \citep{song2015penalized,luckett2019estimating} and machine learning \citep{yang2019sample,jin2020provably}.
For example, the $Q^*$-function of a linear MDP is a linear function of features defined on $\bs$ and $a$ \citep{yang2019sample,jin2020provably,hao2021online}.
In different contexts, $\bs$ can be high-dimensional, such as the feature vector constructed by basis functions \citep{sutton2018reinforcement,luckett2019estimating,jin2020provably} as an approximation to nonparametric $Q^*$-functions. The feature $\phi(\bs)$ can also be taken as a given representation learned from kernel methods or neural networks. 

Accordingly, the task similarity under linear approximation \eqref{eqn:q-k-linear} and \eqref{eqn:delta-k-linear} is defined by the $\ell_q$-sparsity of $\bdelta^{(k)}_t$ for $q\in[0,1]$:
\begin{equation} \label{eq-delta-sparse}
\max_{t \in [T], k\in[K]} \|\bdelta_t^{(k)}\|_q \leq h. 
\end{equation}

Let $\bW_t^{(k)}$ be a ${n_k\times p}$ matrix whose $i$-th row is $\bw(\bs_{i,t}^{(k)}, a_{i,t}^{(k)})$.
In the linear case, we employ the Trans-Lasso algorithm \citep{li2022transfer-jrssb}, a representative of transferred linear regression methods, owing to its simplicity and rate-optimal properties.
Specifically, line 5 - 6 in Algorithm \ref{alg-master} is instantiated with:
\begin{equation*}
\begin{aligned}
\hat{\bb}_t&=\argmin_{\bb\in\RR^p} \left\{\sum_{k=1}^K\|\hat{\by}_t^{(tl-k)}-\bW_t^{(k)}\bb\|_2^2+\lam_{\aux} \|\bb\|_1\right\}.\\
\hat{\bdelta}_t&=\argmin_{\bdelta\in\RR^p} \left\{\|\hat{\by}_t^{(0)}-\bW^{(0)}_t(\bdelta+\hat{\bb}_t)\|_2^2+\lam_0\|\bdelta\|_1\right\}.
\end{aligned}
\end{equation*}
We threshold $\hat{\bdelta}_t$ such that $(\check{\bdelta}_t)_j=(\hat{\bdelta}_t)_j\mathbbm{1}(|(\hat{\bdelta}_t)_j|\geq \lam_0)$ and
threshold $\hat{\bb}_t$ such that $(\check{\bb}_t)_j=(\hat{\bb}_t)_j\mathbbm{1}(|(\hat{\bb}_t)_j|\geq \lam_{\aux})$.
The final output is $\hat{\btheta}^{(0)}_t=\check{\bb}_t+\check{\bdelta}_t$ and $\hat{Q}^{(0)}_t(\bs, a) = [\bphi(\bs)^{\top} \bbone(a=1), \cdots, \bphi(\bs)^{\top}\bbone(a=M)]\hat{\btheta}^{(0)}_t$ for all stage $t \in [T]$.

\begin{algorithm}[t]
\SetKwInOut{Input}{Input}
\SetKwInOut{Output}{Output}
\Input{Target data $\{\{\bs^{(0)}_{t,i}, a^{(0)}_{t,i},r^{(0)}_{t,i}\}_{t\in [T]}\}_{i\in[n_0]}$, \\
source data $\{\{\{\bs^{(k)}_{t,i},a^{(k)}_{t,i},r^{(k)}_{t,i}\}_{t\in[T]}\}_{i\in[n_k]}\}_{k=1}^K$, and \\
discount factor $\gamma\in[0,1]$.}

Construct features $\bW_t^{(k)}$ as a ${n_k\times p}$ matrix whose $i$-th row is $\bw(\bs_{i,t}^{(k)}, a_{i,t}^{(k)})$ for $k\in\{0\}\cup[\cK]$ where $\bw(\bs, a)=[\bphi(\bs)^{\top} \bbone(a=1), \cdots, \bphi(\bs)^{\top} \bbone(a=M)]$. 

Let $\hat\btheta^{(0)}_{T+1}=0$ to deal with pseudo-response construction at last stage $T$.

\For{$t=T,\dots, 1$}{

Construct vectors of pseudo-response $\hat y^{(tl-k)}_{t,i}$ \eqref{eq-mt-pseudo} and $\hat y^{(0)}_{t,i}$ \eqref{yk-pseudo} for the linear setting by
\begin{align*}
\hat{\by}^{(tl-k)}_t & =\br_t^{(k)}+\gamma\bW_t^{(k)}\hat\btheta^{(0)}_{t+1}, \quad k=1,\dots, K\\
\hat{\by}^{(0)}_t & =\br_t^{(0)}+\gamma\bW_t^{(0)}\hat\btheta^{(0)}_{t+1}.
\end{align*}

Apply the transfer learning algorithm for supervised regression:  
\begin{equation*}
\begin{aligned}
\hat{\bb}_t&=\argmin_{\bw\in\RR^p} \left\{\sum_{k=1}^K\|\hat{\by}_t^{(tl-k)}-\bW_t^{(k)}\bb\|_2^2+\lam_{\aux} \|\bb\|_1\right\}.\\
\hat{\bdelta}_t&=\argmin_{\bdelta\in\RR^p} \left\{\|\hat{\by}_t^{(0)}-\bW^{(0)}_t(\bdelta+\hat{\bb}_t)\|_2^2+\lam_0\|\bdelta\|_1\right\}.
\end{aligned}
\end{equation*}
Threshold $  \hat{\bdelta}_t$ such that $(\check{\bdelta}_t)_j=(\hat{\bdelta}_t)_j\mathbbm{1}(|(\hat{\bdelta}_t)_j|\geq \lam_0)$.
Threshold $\hat{\bb}_t$ such that $(\check{\bb}_t)_j=(\hat{\bb}_t)_j\mathbbm{1}(|(\hat{\bb}_t)_j|\geq \lam_{\aux})$.

Calculate $\hat{\btheta}^{(0)}_t=\check{\bb}_t+\check{\bdelta}_t$.
}

\Output{$\{\hat{\btheta}^{(0)}_t\}_{t=[T]}$.}

\caption{Transfer $Q$-learning algorithm (Linear)}
\label{alg-linear}
\end{algorithm}

\subsection{Theoretical Guarantees under Linear Function Approximation}

In this section, we provide rigorous error and regret analyses of both offline and online transfer $Q$-learning under the similarity characterization \eqref{eq-delta-sparse}.
Under such a characterization, the parameter space we consider is
\[
\Omega_q(s,h) = \left\{ \{\btheta^{(0)}_t, \{\bdelta^{(k)}_t\}_{k\in[K]})\}_{t\in [T]}:
\max_{t\in [T]}\|\btheta^{(0)}_t\|_0\leq s, \max_{t\in [T], k \in [K]}\|\bdelta_t^{(k)}\|_q\leq h\right\}.
\]
We write $N_{\aux}=\sum_{k=1}^Kn_k$ as the total number of trajectories from the source tasks.

\begin{assumption} [Conditions on the design]
\label{assume-subgaussian-eigenval}
For $k=0,\dots, K$ and $t=1,\dots,T$, $\bs_{t,i}^{(k)}$ are sub-Gaussian with mean zero and independent across tasks indexed by $k$ and across trajectories indexed by $i$.
Let
$(\bw_{t,i}^{(k)})^{\top} = [\bphi(\bs_{t,i}^{(k)})^{\top}\bbone(a_{t,i}^{(k)}=1),\cdots, \bphi(\bs_{t,i}^{(k)})^{\top}\bbone(a_{t,i}^{(k)}=M)]$ for some known basis functions $\bphi(\cdot)$ and denote the dimension of $\bw_{t,i}^{(k)}$ as $p$. 
The covariance matrices $\bSig^{(k)}_t=\EE[\bw_{t,i}^{(k)}(\bw_{t,i}^{(k)})^{\top}]$ are positive definite with bounded eigenvalues, $t=1,\dots, T$, $k=1,\dots, K$. Moreover,
\begin{align}
\label{Sig-sim}
\max_{t\in [T]}\max_{ k\in [K]} \norm{\bSig^{(k)}_t (\bSig^{(0)}_t)^{-1} - I_{2p}}_{\infty,1}\leq C_{\Sig}<1.
\end{align}
\end{assumption}


Assumption \ref{assume-subgaussian-eigenval} requires sub-Gaussian designs for all the tasks. The positive definiteness of $\bSig_t^{(k)}$ assumes regularity of the covariance of $\bs_{t,i}^{(k)}$ but also requires $a_{t,i}^{(k)}$ has constant variance across samples.
Moreover, we require in (\ref{Sig-sim}) that the covariance matrices for the source and target tasks are sufficiently similar. 

\subsubsection{Convergence rate for the linear setting}
\label{sec4-1}
We provide theoretical guarantees for the offline-to-offline transfer in this subsection.
Theorem \ref{thm1-tl} establishes the convergence rate for $\{\hat{\btheta}^{(0)}_t\}_{t\in[T]}$ in this setting.

\begin{theorem}[Convergence rate of Algorithm \ref{alg-linear}]
\label{thm1-tl}
Suppose that Assumptions \ref{assume-subgaussian-eigenval} 
hold and \eqref{eq-delta-sparse} holds with $q=1$.
Let $N_{\aux}$ be the total number of samples in source tasks.
We take the tuning parameters to be
\[
\lam_{\aux} =
c_1\sqrt{\frac{\log p}{N_{\aux}}}
+ c_1\sqrt{\frac{h}{s}} \paran{\frac{\log p}{n_0}}^{1/4}
~\text{and}~
\lam_0=c_1\sqrt{\frac{\log p}{n_0}}.
\]
Under the sample size condition that $s\sqrt{\log p/N_{\aux}}+h+sh\sqrt{\log p/n_0}\leq C$, for any $t=1,\dots, T$, we have
\begin{equation*}
\frac{1}{n_0} \norm{ \bW^{(0)}_t (\hat{\btheta}^{(0)}_t - \btheta^{(0)}_t) }_2^2\vee\norm{ \hat{\btheta}^{(0)}_t-\btheta^{(0)}_t }_2^2
\lesssim \norm{ \hat{\btheta}^{(0)}_{t+1}-\btheta^{(0)}_{t+1}}_2^2
+ \frac{s\log p}{N_{\aux}}+h\sqrt{\frac{\log p}{n_0}} .
\end{equation*}
with probability at least $1-\exp(-c_2\log p)$.
\end{theorem}

Theorem \ref{thm1-tl} establishes the estimation accuracy of the target $Q^*$-function at each stage.
The rate for the final stage ($t=T$) is the minimax optimal rate in supervised regression with transfer learning \citep{li2022transfer-jrssb} since the final-stage reward equals the value of the state-action pair.
For the $t$-th stage ($t<T$), the estimation error from $\hat{\btheta}^{(0)}_{t+1}-\btheta^{(0)}_{t+1}$ accumulates, as a consequence of using pseudo-responses.
Thus, empirically one may expect that the estimation errors become larger in the earlier stages, even though the convergence rate of each stage are the same.
In the upper bound, the sparsity $s$ of the target parameter is weighted with $1/N_{\aux}$ and the maximum sparsity $h$ of the contrast vectors is weighted with $1/\sqrt{n_0}$.
This shows the improvement of transfer learning, where the $s$-sparse component is learned based on all the studies but the task-specific contrast vectors can only be identified based on the target study with $n_0$ samples. 
When the similarity is sufficiently high, then $h\ll s\sqrt{\log p/n_0}$ and transfer learning can lead to improvements in this case. 
In the theorem, we our choice of $\lam_{\aux}$ depends on the unknown $h/s$, which is for a convenient proof of desirable convergence rate. In practice, $\lam_{\aux}$ can be chosen by cross-validation.

\begin{remark}[Convergence rate of single-task $Q$-learning]
Let $\hat{\btheta}_t^{(st)}$, $t\in[T]$, be the single-task $Q$-learning estimator studied in \cite{song2015penalized}.  For any $t=1,\dots, T$,
\begin{align*}
&\|\hat{\btheta}_t^{(st)}-\btheta^{(0)}_t\|_2^2\lesssim\|\hat{\btheta}^{(st)}_{t+1}-\btheta^{(0)}_{t+1}\|_2^2+\frac{s\log p}{n_0}.
\end{align*}
\end{remark}

For the final stage, the transfer learning estimator $\widehat{\btheta}^{(0)}_T$ has faster convergence rate than the single-task estimator $\hat{\btheta}_t^{(st)}$ as long as $N_{\aux}\gg n_0$ and $h\ll s\sqrt{\log p/n_0}$. Furthermore, the convergence rate of $\widehat{\btheta}^{(0)}_T$ is minimax optimal in the parameter space of interest.
For the $t$-th stage for $t<T$, the gain of transfer learning has two aspects. The first is the error inherited from the next stage, $\|\hat{\btheta}^{(0)}_{t+1}-\btheta^{(0)}_{t+1}\|_2^2$, which is smaller than its single-task counterpart $\|\hat{\btheta}_{t+1}^{(st)}-\btheta^{(0)}_{t+1}\|_2^2$. The second gain comes from aggregating the data at the current stage.

\begin{remark}[Similarity characterization with $q\in[0,1)$]
In Theorem \ref{thm1-tl}, we require $\bdelta_t^{(k)}$ to be approximately sparse ($q=1$). For $q\in[0,1)$, transferred Q-learning algorithms can be analogously developed based on Algorithm 1 in the supplement of \cite{li2022transfer-jrssb}, which is a minimax optimal approach for transfer learning in linear models for $q\in[0,1)$.
\end{remark}

\begin{remark}[Limitations under transition shift]
Our theoretical guarantees rely on stage-wise reward similarity together with mild design similarity across tasks. 
If transition dynamics differ substantially, thereby violating the covariance-similarity condition, or if the reward discrepancies $h$ are large, the advantages of transfer learning may diminish or even disappear. 
These extensions call for new estimation tools and theoretical analysis, but they provide a principled roadmap for expanding our framework beyond the reward-similarity regime studied here. 
A full development of these methods is beyond the scope of this paper and will be pursued in future work.
\end{remark}

\subsubsection{Regret bound of offline-to-online transfer $Q$-learning}
\label{sec-theory-2}
In this subsection, we provide theoretical guarantees for the online Algorithm \ref{alg-online} with knowledge transferred from offline data for $M=2$. Results for a finite $M$ can be derived similarly, with the final outcomes differing only by a constant factor dependent on $M$.
In the online setting, the learner aims to minimize the cumulative regret that measures the expected loss of following the estimated optimal policy instead of the oracle optimal policy.
Mathematically, the cumulative regret over $T$ episodes is defined as
\begin{equation} \label{eqn-cum-regret}
{\rm Regret}_{NT}
=
\sum_{i=1}^N
\sum_{t=1}^T \gamma^t \paran{ \EE[r_{t,i}^{(0)}|\bs^{(0)}_{t,i},a^*_{t,i}]-\EE[r_{t,i}^{(0)}|\bs^{(0)}_{t,i},\hat{a}^{(0)}_{t,i}]},
\end{equation}
where the estimated optimal policy is $\hat{a}_{t,i}^{(0)} = \argmax_{a'\in\{-1,1\}}Q(\bs_{t,i}^{(0)},a';\widehat{\btheta}^{(0)}_t)$
and the oracle optimal policy is $a^*_{t,i} = \argmax_{a'\in\{-1,1\}}Q(\bs_{t,i}^{(0)},a'; \btheta^{(0)}_t)$.
Under the linear $Q^*$-function \eqref{eqn:q-k-linear} with $M=2$, they can be further simplified to
\[
\hat{a}_{t,i}^{(0)} = sgn((\bs^{(0)}_{t,i})^{\top}\widehat{\bpsi}_t),
\quad\text{and}\quad
a^*_{t,i}=sgn((\bs^{(0)}_{t,i})^{\top}\bpsi_t),
\]
where $sgn(\cdot)$ is the sign function.
The regret bound of online learning with offline transfer is given in Theorem \ref{thm-ol}.

\begin{theorem}[Cumulative regret of Algorithm \ref{alg-online} after $T$ rounds]
\label{thm-ol}
Suppose that Assumptions \ref{assume-subgaussian-eigenval}
hold, (\ref{eq-delta-sparse}) holds with $q=1$. and $s\sqrt{\log p/N_{\aux}}+h\leq C$, $n_e\gtrsim s^2h^2\log p+\log p$.
We take tuning parameters to be
\[
\lam_{\aux}
=\sqrt{\frac{\log p}{N_{\aux}}}
+ \sqrt{\frac{h}{s}} \paran{ \frac{\log p}{n_e}}^{1/4}
~~\text{and}~~
\lam_0=\sqrt{\frac{\log p}{n_e}}.
\]
For any $N>n_e$,
\begin{align}
\label{reg-tl}
{\rm Regret}_{NT}
\lesssim
\frac{n_e\gamma }{1-\gamma}
+
(N-n_e) \paran{\sqrt{\frac{s\log p}{N_{\aux}}}+ h^{1/2} \paran{\frac{\log p}{n_e}}^{1/4}}
\end{align}
with probability at least $1-\exp(-c_1\log p)$.
\end{theorem}
We now find the optimal choice of $n_e$, which minimizes the RHS of (\ref{reg-tl}). To simplify the analysis, we parameterize $h=N_{\aux}^{-\alpha}$ for some $\alpha\geq 0$.
In the supplementary files, we show that if we take
\begin{align*}
n_e \asymp   \max\left\{\frac{N^{4/5}(\log p)^{1/5}}{N_{\aux}^{2\alpha/5}},\frac{s^2\log p}{N_{\aux}^{2\alpha}}\right\},
\end{align*}
then with probability at least $1-\exp(-c_1\log p)$
\begin{equation}
\label{reg-tl-opt}
{\rm Regret}_{NT}
\lesssim  \max\left\{\frac{N^{4/5}(\log p)^{1/5}}{N_{\aux}^{2\alpha/5}},\frac{s^2\log p}{N_{\aux}^{2\alpha}}\right\}+N\sqrt{\frac{s\log p}{N_{\aux}}}.
\end{equation}
We see that the larger the $N_{\aux}$, i.e., more source data, the smaller the cumulative regret; and the larger the value of $\alpha$, i.e., higher the similarity, the smaller the cumulative regret.

\begin{remark}[Cumulative regret of single-task $Q$-learning]
\label{re3}
Without offline data, we denote the estimated optimal policy by $\widehat{a}^{(st)}$. It is easy to show that for $n_e\gg (s\log p)^2$, then with probability at least $1-\exp(-c_1\log p)$,
\begin{align}
\label{reg-st}
{\rm Regret}_{NT}^{(st)}
\lesssim
\frac{n_e\gamma}{1-\gamma}+(N-n_e)\sqrt{\frac{s\log p}{n_e}}.
\end{align}
with probability at least $1-\exp(-c_1\log p)$. If we take $n_e\asymp N^{2/3}(s\log p)^{1/3}\vee (s\log p)^2$, then
\begin{align}
\label{reg-st-opt}
{\rm Regret}_{NT}^{(st)}\lesssim N^{2/3}(s\log p)^{1/3}+ (s\log p)^2,
\end{align} with probability at least $1-\exp(-c_1\log p)$.
\end{remark}
Comparing \eqref{reg-tl} with \eqref{reg-st}, we see that the cumulative regret of transfer $Q$-learning is always smaller when $h=o(s\sqrt{\log p/n_e})$.
Comparing \eqref{reg-tl-opt} with \eqref{reg-st-opt}, we see that as long as $N_{\aux}>N^{1/(3\alpha)}+N^{2/3}$, the regret of transfer learning policy is always no larger than the single-task policy. This comparison further implies that if $N$ is very large, then the optimal choice of $n_e$ can be larger than $N_{\aux}$. In this case, transfer learning may not yield a significant improvement and it suffices to consider single-task $Q$-learning.

\section{Numerical Studies}
\label{sec5}
In this section, we demonstrate the advantages of the proposed transferred Q-learning algorithm on synthetic and real data sets. Code and implementation details for this section are available online.

\subsection{Two-stage MDP with analytical optimal $Q^*$ function}   \label{sec:}

We first consider a simple MDP model which has an analytical form for the optimal $Q^*$ function.
In such a setting we can explicitly compare the estimated $Q^*$ function with the ground truth.
The generative model is designed based on those in \cite{chakraborty2010inference} and \cite{song2015penalized}.
The underlying model is a two-stage MDP with two possible actions $\calA=\braces{-1,1}$ and two states $\calS = \braces{-1,1}$.
The binary states $S_t$ and the binary treatment $A_t$ are generated as follows:
\begin{equation*}
\begin{aligned}
&\Pr\paran{S_1=-1} = \Pr\paran{S_0=1} = 0.5, \\
&\Pr\paran{A_t=-1} = \Pr\paran{A_t=1} = 0.5, \quad t = 1, 2, \\
&\Pr\paran{S_2 | S_1, A_1} = 1 - \Pr\paran{S_2 | S_1, A_1} = {\rm expit}\paran{b_1 S_1 + b_2 A_1},
\end{aligned}
\end{equation*}
where ${\rm expit}\paran{x} = \exp\paran{x} / \paran{1 + \exp\paran{x}}$.
The immediate reward $R_1 = 0$ and $R_2$ is given by
\begin{equation*}
R_2 = \kappa_1 + \kappa_2 S_1 + \kappa_3 A_1 + \kappa_4 S_1 A_1 + \kappa_5 A_2 + \kappa_6 S_2 A_2 + \kappa_7 A_1 A_2 + \eps_2,
\end{equation*}
where $\eps_2 \sim \calN\paran{0, 1}$.
Under this setting, the true $Q^*$ functions for time $t=1, 2$ are
\begin{equation}  \label{eqn:true-q}
\begin{aligned}
Q_2\paran{S_2, A_2; \btheta_2} & = \theta_{2,1} + \theta_{2,2} S_1 + \theta_{2,3} A_1 + \theta_{2,4} S_1 A_1 \\
& ~~~ + \theta_{2,5} A_2 + \theta_{2,6} S_2 A_2 + \theta_{2,7} A_1 A_2 \\
Q_1\paran{S_1, A_1; \btheta_1} & = \theta_{1,1} + \theta_{1,2} S_1 + \theta_{1,3} A_1 + \theta_{1,4} S_1 A_1,
\end{aligned}
\end{equation}
where the true coefficients $\btheta_t$ are explicitly functions of $b_1$, $b_2$, $\kappa_1,\dots, \kappa_7$ given in \eqref{eqn:true-q-theta} in the supplemental material.
At each state, the observed covariates $X_t\in \RR^p$, $p=100$, consist of first eight elements $1$, $S_1$, $A_1$, $S_1A_1$, $A_2$, $S_2$, $S_2 A_2$, $A_1 A_2$ and the remaining elements that are randomly sampled from standard normal.

We consider transfer between two different but similar MDPs from the above model.
The target and source MDPs are different in the coefficients $\kappa$'s and therefore $\theta$'s in \eqref{eqn:true-q}.
Specifically, we set $\theta_{2,j} = 1$, $1 \le j \le 7$ for the target MDP.
The second-stage coefficients of the source task are the same except that the second element $\theta_{2,2}^{(1)}  = 1.2$.
Therefore, according to equation \eqref{eqn:true-q-theta} in the supplemental material, the true coefficients of stage-one $Q^*$ functions are $\theta_{1,1}, \theta_{1,2}, \theta_{1,3},  \theta_{1,4} \approx
2.69, 1.19, 1.69, 1.19$ for the target MDP, and $\theta_{1,1}^{(1)}, \theta_{1,2}^{(1)}, \theta_{1,3}^{(1)},  \theta_{1,4}^{(1)}  \approx
2.69, 1.39, 1.69, 1.19$ for the source MDP.
In summary, the true $Q_1$ functions of the target and source MDPs are different only in $\theta_{1,2}$ and the $Q_2$ functions are different only in $\theta_{2,2}$.
The coefficients of the remaining elements in $X_t$ are all set to zero.
We generate trajectories of the form $\paran{\bx_{1,i}, a_{1,i}, r_{1,i}, \bx_{2,i}, a_{2,i}, r_{2,i}}$ from both target and source MDPs.
The target task consists of $n_0$ trajectories while the source task consists of $n_1$ trajectories.

\begin{figure}[t]
\centering
\includegraphics[width=\linewidth]{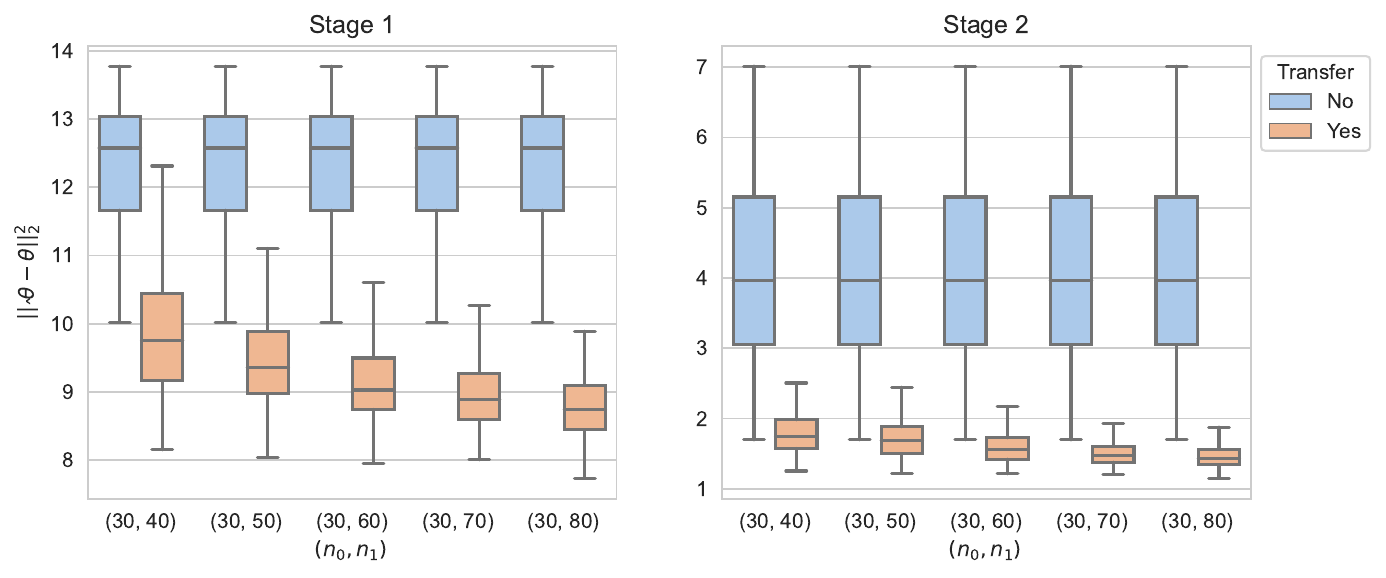}
\includegraphics[width=\linewidth]{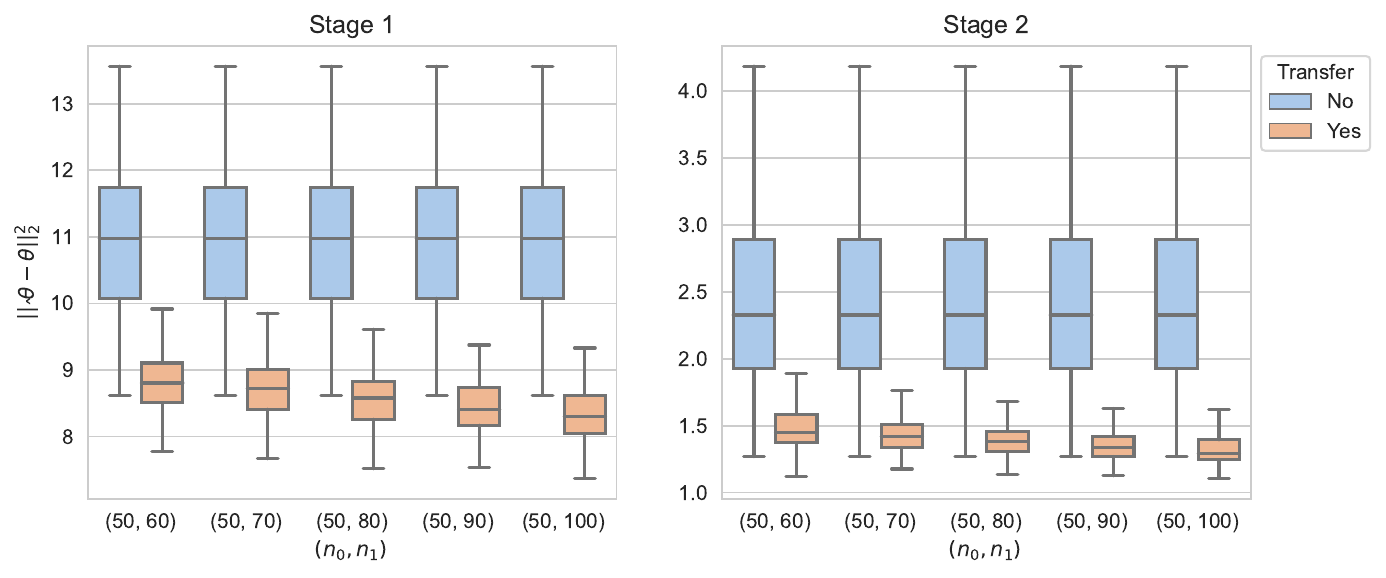}
\caption{\label{fig:chakra-coef}Comparison of the estimated coefficients of the optimal $Q^*$ function.
The $y$-axis represents $\norm{\hat\btheta_t - \btheta_t}_2^2$ for $t=1$ (left) and $t=2$ (right).
The dimension is $p =100$ and sparsity $s = 7$.
} 
\end{figure}

\begin{figure}[t]
\centering
\includegraphics[width=\linewidth]{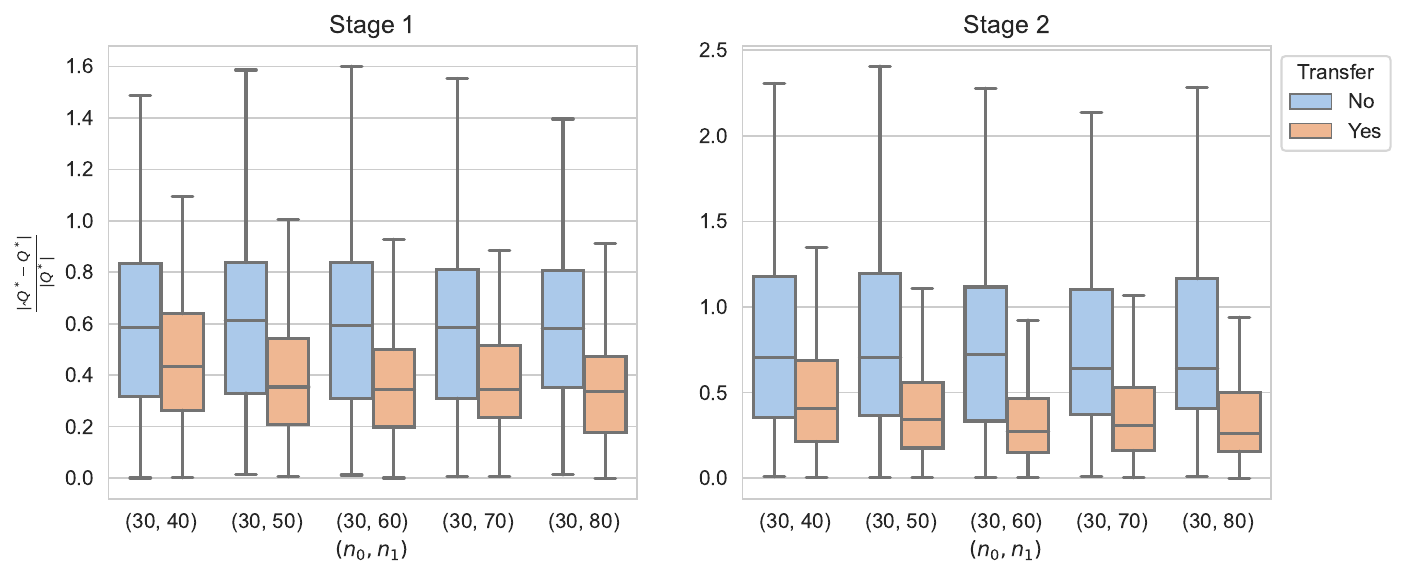}
\includegraphics[width=\linewidth]{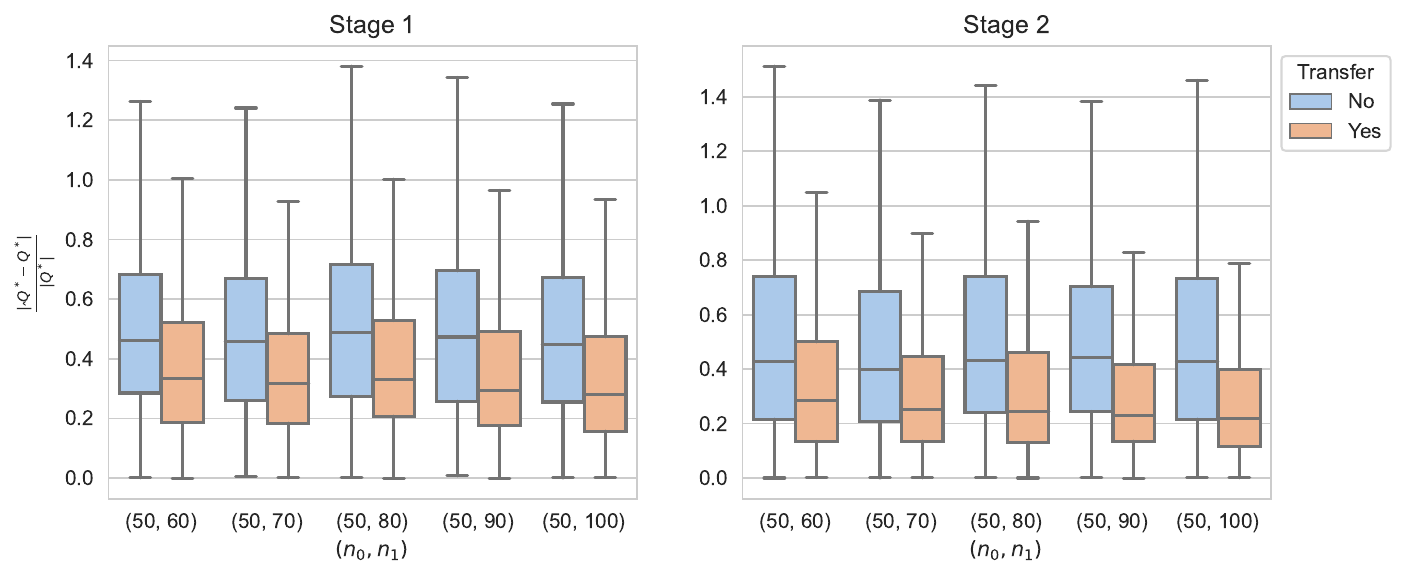}
\caption{\label{fig:chakra-pred}Comparison of the prediction of the optimal $Q^*$ function with different sample size configurations.
The $y$-axis represents $\abs{\hat Q^* - Q^*}/\abs{Q^*}$.
The dimension is $p =100$ and sparsity $s = 7$.}
\end{figure}

We compare $\norm{\hat\btheta_t - \btheta_t}_2^2$, $t=1,2$, with or without transfer under combinations of $n_0$ and $n_{\aux}=n_1$.
The boxplots are shown in Figure \ref{fig:chakra-coef}.
We also generate a new dataset and compare predicted $\hat Q^*(\bx, a)$ obtained by transferred $Q^*$ learning and its vanilla counterpart.
The boxplots of $\abs{\hat Q^*(\bx, a) - Q^*(\bx, a)}/\abs{Q^*(\bx, a)}$ (averaged over all state-action pairs in the new dataset) are presented in Figure \ref{fig:chakra-pred}.
It is clear that transfer $Q$-learning performs much better than the vanilla $Q$-learning without transfer in terms of both coefficient estimation and prediction.
The advantage is more prominent in earlier stages since the transfer benefit in the latter stage positively cascade to the earlier stages through the second term in \eqref{eq-mt}.

Table \ref{tab:opt-act} compares the frequency of correct optimal actions chosen by single-task $Q$-learning and transfer $Q$-learning with the new dataeset.
We observe that transfer $Q$-learning achieves higher accuracy in choosing the optimal actions across all combinations of $n_0$ and $n_1$, which are the number of trajectories of the target task and the source task, respectively. 
The amplitude of accuracy increase is higher when $n_0$ is small. 
This again shows the advantage of transfer $Q$-learning, especially when the number of trajectories of the target task is small. 

\begin{table}[t]
\centering
\caption{\label{tab:opt-act}  The frequency of correct optimal actions. 
The comparison is between the optimal action $A^{*,(st)}$ chosen by single-task $Q$-learning and the optimal action $A^{*,(tl)}$ chosen by transfer $Q$-learning. 
We consider different combinations of $n_0 = 30, 50, 70$ and $n_1 = n_0 + 10, \, \cdots, n_0 + 50$, where $n_0$ and $n_1$ denote the number of trajectories of the target task and the source task, respectively. 
} 
{
\begin{tabular}{c|c|ccccc}
\hline
\multirow{2}{*}{{$n_0$}}  & \multirow{2}{*}{{$A^{*,(st)}$}}  &  \multicolumn{5}{c}{ {$A^{*,(tl)}$ with different $n_1$} } \\
\cline{3-7}
&  & $n_0 + 10$ & $n_0 + 20$ & $n_0 + 30$ & $n_0 + 40$ & $n_0 + 50$ \\
\hline
30 & 0.555 & 0.965 & 0.985 & 0.990 & 0.970 & 0.985 \\
50  &  0.860  & 0.990 &  0.995 & 1.000 & 1.000 & 1.000 \\
70 & 0.905 & 1 & 1 & 1 & 1 & 1 \\
\hline
\end{tabular}
}
\end{table}

\subsection{Offline-to-online transfer}

In this section, we study the empirical performance of online transfer $Q$-learning using an offline source dataset.
The generative MDP is the same as that defined in \eqref{eqn:true-q}.
We have access to an offline source dataset of trajectories $\paran{X_1, A_1, R_1, X_2, A_2, R_2}$ generated from an MDP with $\theta_{2j}^{(1)} = \kappa_j^{(1)} = 1$, $1 \le j \le 3$,  $5 \le j \le 7$ and $\theta_{2j}^{(1)} = \kappa_j^{(1)} = 2$, $j=4$.
The online target RL task is modeled by an MDP with $\theta_{2j}^{(1)} = \kappa_j^{(1)} = 1$, $1 \le j \le 7$. 
The dimension of covariates is $p = 100$.

\begin{figure}[t]
\centering
\includegraphics[width=0.48\linewidth,height=5cm]{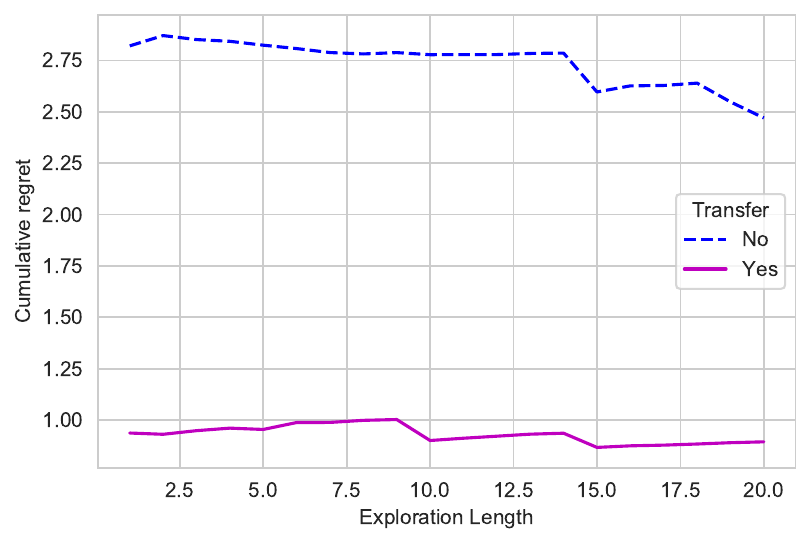}
\includegraphics[width=0.48\linewidth,height=5cm]{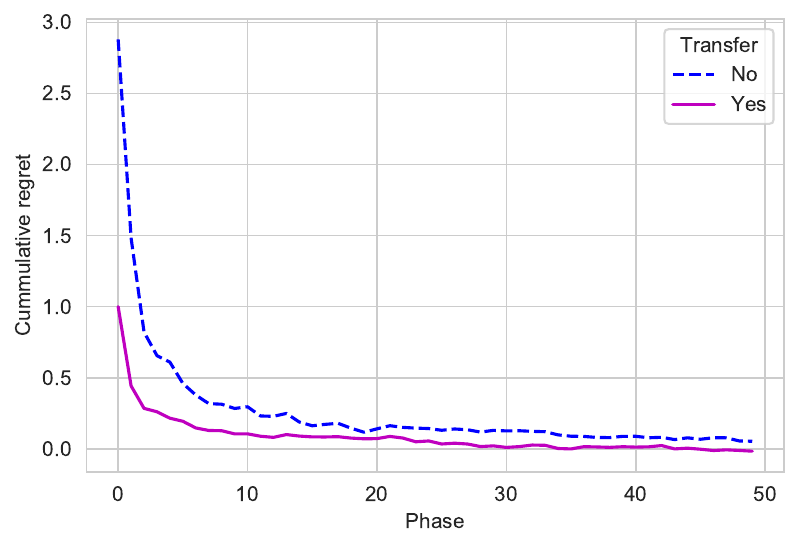}
\caption{\label{fig:chakra-online}
Cumulative regret of online $Q$-learning with or without transfer. 
Left panel: The  {\em explore-then-commit} (ETC) online $Q$-learning; 
Right panel: phase-based ETC online $Q$-learning with parameter updates.
} 
\end{figure}

We first study the cumulative regret of transferred and vanilla $Q$-learning with different lengths of exploration, $n_{e} \in \braces{1, 2, \cdots, 20}$.
The size of the offline source dataset is $n_1= 100$.
At the exploration stage, $n_{e}$ trajectories are generated by random actions.
The coefficients $\hat\btheta_t$ are estimated using $Q$-learning with or without transfer.
The average cumulative regret at the exploitation stage versus different length of exploration are presented in Figure \ref{fig:chakra-online} (a).
The length of exploitation stage is $100$.
For each $n_{e}$, the mean of the cumulative regret at  exploitation stage is reported since the values of $\hat\btheta_t^{(0)}$ are not updated during exploitation stage.
We observe that under the condition that $h\ll s\sqrt{\log p/n_{e}}$, the regret of transfer $Q$-learning is much smaller than that of vanilla $Q$-learning, which is consistent with the result in Theorem \ref{thm-ol}.
Since Algorithm \ref{alg-online} is of the {\em explore-then-commit} (ETC) type, the advantage of transferred $Q^*$ learning shown in the left panel of Figure \ref{fig:chakra-online} can be viewed as the jumpstart improvement---one of the three main objectives of transfer learning defined in \cite{langley2006transfer} and  \cite{taylor2009transfer}.

We also empirically study the cumulative regret of online $Q$-learning where the values of $\hat\btheta_t$, $t=1,2$ are updated during the exploitation stage.
This phase-based ETC online $Q$-learning algorithm is a natural extension of Algorithm \ref{alg-online} and it goes as follows.
At the first phase, vanilla $Q$-learning initializes $\hat\btheta_t$ to zero, while transfer $Q$-learning initializes with $\hat\btheta_t$ that are estimated using offline trajectories from the source task.
Then at each phase, a batch of $100$ trajectories are generated using greedy actions based on estimated $\hat\btheta_t$.
Using the extra generated trajectories at the current phase, the values $\hat\btheta_t$ are updated and will be used to generate greedy actions for the next phase.
The right panel of Figure \ref{fig:chakra-online} shows the mean of cumulative regret as phases proceed online.
It shows all three main advantages of transferred learning, namely, jumpstart, learning speed, and asymptotic improvements \citep{langley2006transfer,taylor2009transfer}.

\subsection{Medical data application: {\sc Mimic-iii} Sepsis cohort} \label{sec:appl-mimic}

In this section, we illustrate an application of the transfer $Q$-learning in the Medical Information Mart for Intensive Care version III (MIMIC-III) Database \citep{johnson2016mimic}, which is a freely available source of de-identified critical care data from 2001 -- 2012 in six ICUs at a Boston teaching hospital.

We consider a cohort of sepsis patients, following the same data processing procedure as that in \cite{chen2022reinforcement,komorowski2018artificial}.
Each patient in the cohort is characterized by a set of 47 variables, including demographics, Elixhauser premorbid status, vital signs, and laboratory values.
Patients' data were coded as multidimensional discrete time series $\bx_{i,t} \in\RR^{47}$ for $1\le i \le N$ and $1\le t \le T_i$ with 4-hour time steps.
The actions of interests are the total volume of intravenous (IV) fluids and maximum dose of vasopressors administrated over each 4-hour period.
We discretize two actions into three levels (i.e., low, medium, and high), respectively.
In our setting, the low-level corresponds to level 1 - 2, the medium-level corresponds to level 3 and the high-level corresponds to level 4 - 5 in \cite{komorowski2018artificial}.
The combination of the two drugs makes $M = 3 \times 3 = 9$ possible actions in total.
The final processed dataset contains 20943 unique adult ICU admissions, among which 11704 (55.88\%) are female (0) and 9239 (44.11\%) are male (1).

The reward signal is important and is crafted carefully in real applications.
For the final reward, we follow \cite{komorowski2018artificial} and use hospital mortality or 90-day mortality.
Specifically, when a patient survived after 90 days out of hospital, a positive reward was released at the end of each patient's trajectory; a negative reward was issued if the patient died in hospital or within 90 days out of hospital.
In our dataset, the mortality rate is 24.21\% for female and 22.71\% for male.
For the intermediate rewards, we follow \cite{prasad2017reinforcement} and associates reward to the health measurement of a patient.
The detailed description of the data pre-processing is presented in Section \ref{appen:mimic-iii} of the supplemental material.

\begin{figure}[t]
\centering
\includegraphics[width=0.5\linewidth]{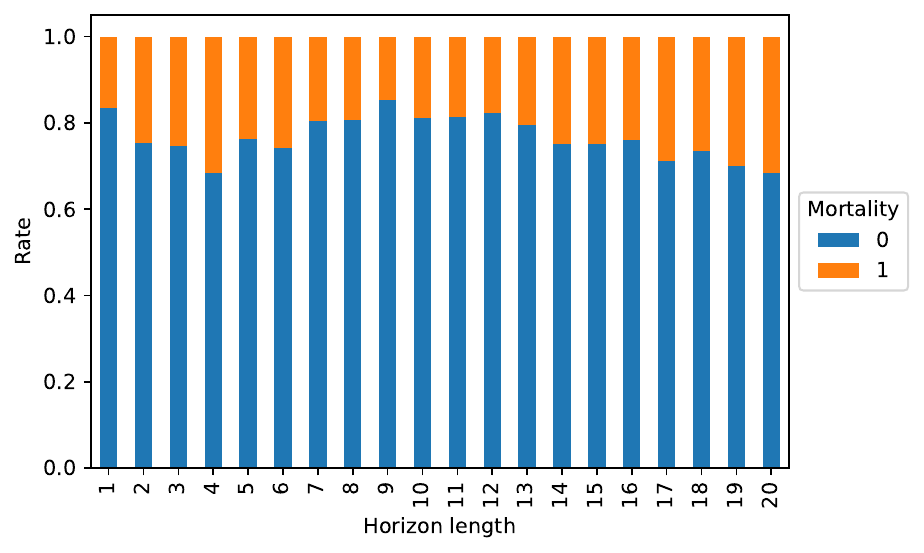}
\caption{\label{fig:mort-horizon}Mortality rates across different horizon lengths.} 
\end{figure}

The trajectory horizons are different in the dataset, with the maximum being 20 and minimum being 1.
We use 20 as the total length of horizon.
The trajectories are aligned at the last steps while allowing the starting steps vary.
For examples, the trajectories with length 20 start at step $1$ while the trajectories with length 10 start at step $11$.
But they all end at step 20.
We adopt  this method because the distribution of final status are similar across trajectories.
Figure \ref{fig:mort-horizon} presents the mortality rates of different lengths.
We see that while the numbers of trajectories differ a lot, the mortality rates do not vary much across trajectories with different horizons.
On the contrary, the starting status of patients may be very different. 
The one with trajectory length $20$ may be in a worse status and needs $10$ steps to reach the status similar to the starting status of the one with length $10$. 
We believe this is a reasonable setup to illustrate our method.
A rigorous medical analysis is beyond the scope of this paper and is a worthwhile topic for future research.

We consider transfer $Q$-learning across genders.
The analytical model for optimal $Q_t^{(k)}$ function for each step $t  \in [20]$ is modeled by
\begin{equation} \label{eqn:anal-opt-Q}
\begin{aligned}
Q_t^{(k)}\paran{\bx, a} \approx \bx^\top \sum_{a'=1}^9 \btheta_{t,a}^{(k)} \bone\paran{a = a'},
\end{aligned}
\end{equation}
where the covariates $\bx$ contains 44 variables detailed in Table \ref{tab:variables} in the supplements.
Even though the total number of trajectories of gender $0$ is large, estimating \eqref{eqn:anal-opt-Q} is still a high-dimensional problem since we allow $\btheta_{t,a}^{(k)}$ be different across step $t$, action $a$ and gender subgroup $k = 0,1$ and the number of trajectories corresponding to a specific combination of $t$, $a$, and $g$ is small. 
For example, there are only 19 samples available to estimate $\btheta_{t,a}^{(k)}$ for gender $k = 1$, step $t = 1$, and action $a$ corresponding to the combination $(\textup{IV}, \textup{Vaso}) = (0, 1)$.
Table \ref{tab:least} in the supplemental material shows that the least ten samples sizes are all under $30$.
We observe that gender $k = 1$ (male) has fewer samples so we consider transfer $Q$-learning with target task for gender $k=1$ and auxiliary source task from gender $k=0$. 

\begin{figure}[t]
\centering
\includegraphics[width=0.8\linewidth, height=5cm]{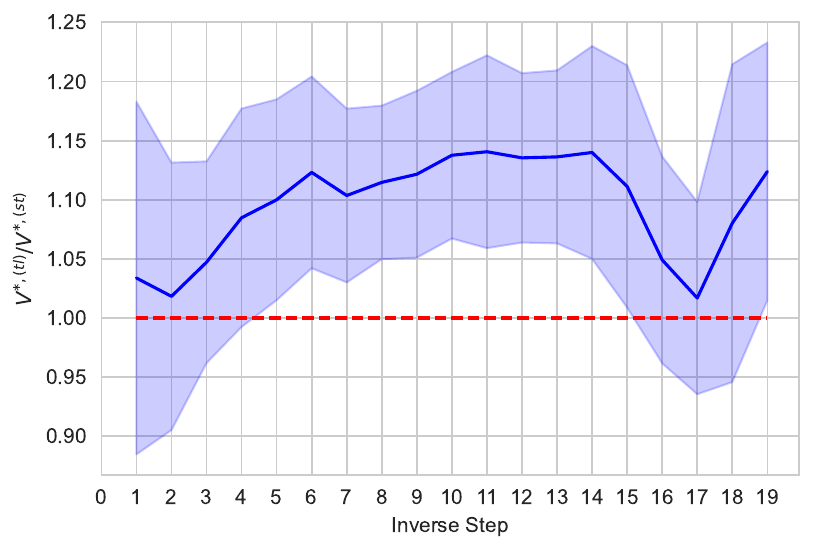}
\caption{\label{fig:optimal-value-ratio} Ratio between optimal aggregated values $V^*$ of transfer $Q$-learning ($V^{*,(tl)}$) and those of Vanilla $Q$-learning ($V^{*,(st)}$) applied on the target data.
Shaded area covers one standard deviation from the mean.
Inverse step is $20 - t$ where $1 \le t \le 20$ is the natural step in a trajectory. } 
\end{figure}

The estimation procedure of transfer $Q$-learning follows Algorithm \ref{alg-master}. 
We set the discount factor as $\gamma = 0.98$. 
We also estimate the $Q^*$ function by the vanilla $Q$-learning which is the counterpart of Algorithm \ref{alg-master} without transfer. 
The Lasso tuning parameters are chosen according to Theorem \ref{thm1-tl} and by a linear search for the value of $c_1$ that maximizes the objective function. 
We calculate the optimal aggregated values of transferred and Vanilla $Q$-learning, denoted by $V^{*,(tl)}$ and $V^{*,(st)}$, respectively. 
Figure \ref{fig:optimal-value-ratio} plot the average of the ratio $V^{*,(tl)} / V^{*,(st)}$ and its standard deviation. 
The mean ratio is above one, indicating that the optimal value of transfer $Q$-learning is larger.

\section{Conclusions}
\label{sec6}

In this work, we envisage $Q$-learning with knowledge transfer in the form of an algorithm that exploits the availability of samples from different but similar RL tasks.
The {\em similarity} between target and source RL tasks is characterized by their corresponding reward functions which can be checked directly in practice.
We note that a salient feature of RL is its multi-stage learning and, accordingly, we design a novel {\em re-targeting} step to enable ``cross-stage transfer'' along multiple stages in an RL task, in addition to the usual ``cross-task transfer'' in TL for supervised learning.
For both offline and online RL tasks, our approach demonstrates improvements over single-task $Q$-learning, both theoretically and empirically.

The similarity definition among multi-stage RL tasks can be diverse and more flexible. For instance, some related tasks may have different action spaces or different number of stages. It is also important to study transfer learning for such situations. 
As the similarity level is always unknown, it is likely that we include some source tasks which are far away from the target task. As a future research direction, one can perform a model selection aggregation step and aggregation in multi-stage models, especially the online setting, is largely an open problem and requires further study.

\begin{funding}
Elynn Chen was supported by NSF Grant DMS-2412577.
\end{funding}



\bibliographystyle{imsart-number} 
\bibliography{main,rl-appl}       

\newpage
\begin{supplement}
	\stitle{Supplementary Material of ``\TITLE''}
	\sdescription{The supplementary material provides detailed mathematical proofs for the main theoretical results, including the convergence rate and regret bounds, together with additional lemmas and intermediate derivations. It also contains implementation details and extended results from the numerical studies, such as full simulation settings, additional tables, and figures. For the real-data application on the MIMIC-III sepsis cohort, the supplement describes the data preprocessing procedures, variable definitions, reward construction, and cohort statistics. These materials ensure full reproducibility and transparency of the theoretical and empirical findings reported in the main paper.}
\end{supplement}
\begin{appendix}
	

\section{Proofs} \label{sec:proof}

We first define some notation. Let $\alpha_k=n_k/N_{\aux}$ and $\bar{\bSig}_t=\sum_{k=1}^K\alpha_k\bSig^{(k)}_t$.
\subsection{Proof of Theorem \ref{thm1-tl}}
Notice that
\[
   \EE[\sum_{k=1}^K(\bW^{(k)}_T)^\top(\by^{(k)}_T-\bW^{(k)}_T\bb_T)]=0
\]
for
\begin{equation}
\label{eq-bt}
    \bb_T=\btheta_t+\underbrace{\{\sum_{k=1}^K\alpha_k\bSig^{(k)}_T\}^{-1}\sum_{k=1}^K\alpha_k\bSig^{(k)}_T\bdelta^{(k)}_T}_{\bar{\bdelta}_T}.
\end{equation}

Define an event
\begin{align*}
\mathcal{E}_t&=\left\{\frac{1}{n_0}\|(\bW_t^{(0)})^\top(\by_t^{(0)}-\bW_t^{(0)}\btheta_t)\|_{\infty}\leq \lam_0/2,\right.\\
&\frac{1}{N_{\aux}}\|\sum_{k=1}^K(\bW_t^{(k)})^\top(\by_t^{(tl-k)}-\bW_t^{(k)}\bb_t)\|_{\infty}\leq \lam_{\aux}/2,\\
&\quad\quad\bu^\top\widehat{\bSig}^{(0)}_t\bu\geq C\|\bu\|_2^2-\|\bu\|_1^2\frac{\log p}{n_0},~\bu^\top\widehat{\bar{\bSig}}_t\bu\geq C\|\bu\|_2^2-\|\bu\|_1^2\frac{\log p}{N_{\aux}},\\
&\quad\quad\left. \sup_{\|\bu\|_0\leq C(s+h/\sqrt{\log p/n_0})} \bu^\top\widehat{\bar{\bSig}}_t\bu\leq C\|\bu\|_2^2\right\}.
\end{align*}
It is easy to show that for $s\log p/N_{\aux}+h\sqrt{\log p/n_0}=o(1)$, $\mathbb{P}(\cap_{t=1}^T\mathcal{E}_t)\geq 1-\exp(-c_1\log p)$ for any finite $T$.

Conceptually, define $\tilde{\bdelta}_T$ be a thresholded version of $\bar{\bdelta}_T$ at threshold level $\sqrt{\log p/n_0}$. That is, $(\tilde{\bdelta}_T)_{j}=(\bar{\bdelta}_T)_j\mathbbm{1}(|(\bar{\bdelta}_T)_j|\geq \sqrt{\log p/n_0})$.
We only leverage $\tilde{\bdelta}_T$ to facilitate the proof.
\begin{lemma}
\label{lem1-tl}
Under the conditions of Theorem \ref{thm1-tl}, with probability at least $1-\exp(-c_1\log p)$ for some constant $c_1>0$,
\begin{align*}
 & \|\hat{\btheta}_T-\btheta_T\|_2^2\lesssim \frac{s\log p}{N_{\aux}}+h\sqrt{\frac{\log p}{n_0}}\wedge h^2\\
  &\|\hat{\btheta}_T+(\bar{\bdelta}_T-\tilde{\bdelta}_T)-\btheta_T\|_0\lesssim |S_T|\lesssim (s+h/\sqrt{\log p/n_0}).
\end{align*}
\end{lemma}
\begin{proof}[Proof of Lemma \ref{lem1-tl}]
For the final stage, the analysis is analogous to the supervised linear regression.
Specifically,
\begin{align*}
  \|\bar{\bdelta}_T\|_1&= \|\bar{\bSig}_T^{-1}\sum_{k=1}^K\alpha_k\bSig^{(k)}_T\bdelta_T^{(k)}\|_1 \\
  &\leq \|\{\bSig_T^{(0)}\}^{-1}\sum_{k=1}^K\alpha_k\bSig^{(k)}_T\bdelta_T^{(k)}\|_1 \\
  & +\|\{\bSig_T^{(0)}\}^{-1}\sum_{k=1}^K\alpha_k(\bSig^{(k)}_T-\bSig^{(0)}_T)\bar{\bSig}_T^{-1}\sum_{k=1}^K\alpha_k\bSig^{(k)}_T\bdelta^{(k)}_T\|_1\\
  &\leq (1+C_{\Sig})\max_{k\leq K}\|\bdelta^{(k)}_T\|_1+C_{\bSig}  \|\bar{\bdelta}_T\|_1.
\end{align*}
Hence, for $C_{\Sig}<1$,
\[
   \|\bar{\bdelta}_T\|_1\leq \frac{1+C_{\Sig}}{1-C_{\Sig}}\max_{k\leq K}\|\delta^{(k)}_T\|_1.
\]
Oracle inequality for $\hat{\bb}_T$:
\begin{align*}
\frac{1}{N_{\aux}}\sum_{k=1}^K\|\bW_T^{(k)}(\hat{\bb}_T-\bb_T)\|_2^2&\leq  \frac{1}{N_{\aux}}|\sum_{k=1}^K\langle \bW_T^{(k)}(\hat{\bb}_T-\bb), \by_T^{(tl-k)}-\bW_T^{(k)}\bb_T)\rangle| \\
& +\lambda_{\aux}(\|\bb_T\|_1-\|\hat{\bb}_T\|_1).
\end{align*}

Using standard arguments, in event $\mathcal{E}_T$,
\begin{equation}
    \begin{aligned}
  &\|\widehat{\bar{\bSig}}_T^{1/2}(\hat{\bb}_T-\bb_T)\|_2^2\vee\|\hat{\bb}_T-\bb_T\|_2^2\leq Cs\lam_{\aux}^2+h\lam_{\aux}\wedge h^2\\
  &\|\hat{\bb}_T-\bb_T\|_1\leq s\lam_{\aux}+h.
  \end{aligned}
  \label{bT-l1}
\end{equation}

For $\hat{\bdelta}_T$, we have the following oracle inequality
\begin{align*}
\frac{1}{n_0}\|\bW^{(0)}_T(\hat{\bdelta}_T-\bar{\bdelta}_T)\|_2^2
& \leq \frac{1}{n_0}\langle \bW^{(0)}_T(\hat{\bdelta}_T-\bar{\bdelta}_T),\br_T^{(0)}-\bW_T^{(0)}(\hat{\bb}_T+\bar{\bdelta}_T)\rangle \\
& +\lam_0\|\bar{\bdelta}_T\|_1-\lam_0\|\hat{\bdelta}_T\|_1\\
&\leq |\frac{1}{n_0}\langle \bW^{(0)}_T(\hat{\bdelta}_T-\bar{\bdelta}_T),\br_T^{(0)}-\bW_T^{(0)}\btheta_T\rangle| \\
& +\underbrace{(\hat{\bdelta}_T-\bar{\bdelta}_T)^\top\widehat{\bSig}^{(0)}_T(\hat{\bb}_T-\bb_T)}_{E_{2,T}}+\lam_0\|\bar{\bdelta}_T\|_1-\lam_0\|\hat{\bdelta}_T\|_1.
\end{align*}
For $E_{2,T}$, we have with probability at least $1-\exp(-c_1\log p)$,
\begin{align*}
&E_{2,T}\leq (\hat{\bdelta}_T-\bar{\bdelta}_T)^\top\bSig^{(0)}_T(\hat{\bb}_T-\bb_T) +|(\hat{\bdelta}_T-\bar{\bdelta}_T)^\top(\widehat{\bSig}^{(0)}_T-\bSig^{(0)}_T)(\hat{\bb}_T-\bb_T)|\\
&= (\hat{\bdelta}_T-\bar{\bdelta}_T)^\top\bSig^{(0)}_T\bar{\bSig}_T^{-1}\bar{\bSig}_T(\hat{\bb}_T-\bb_T)+|(\hat{\bdelta}_T-\bar{\bdelta}_T)^\top(\widehat{\bSig}^{(0)}_T-\bSig^{(0)}_T)(\hat{\bb}_T-\bb_T)|\\
&\leq (\hat{\bdelta}_T-\bar{\bdelta}_T)^\top\bSig^{(0)}_T\bar{\bSig}_T^{-1}\widehat{\bar{\bSig}}_T(\hat{\bb}_T-\bb_T)+|(\hat{\bdelta}_T-\bar{\bdelta}_T)^\top(\widehat{\bSig}^{(0)}_T-\bSig^{(0)}_T)(\hat{\bb}_T-\bb_T)|\\
&\quad +|(\hat{\bdelta}_T-\bar{\bdelta}_T)^\top\bSig^{(0)}_T\bar{\bSig}_T^{-1}(\widehat{\bar{\bSig}}_T-\bar{\bSig}_T)(\hat{\bb}_T-\bb_T)|\\
&\lesssim \|\hat{\bdelta}_T-\bar{\bdelta}_T\|_1 \|\bSig^{(0)}_T\bar{\bSig}_T^{-1}\|_{\infty,1}\lam_{\aux}+\|\hat{\bdelta}_T-\bar{\bdelta}_T\|_1\|\hat{\bb}_T-\bb_T\|_1\sqrt{\frac{\log p}{n_0}}\\
&\quad+ \|\bSig^{(0)}_T\bar{\bSig}_T^{-1}\|_{\infty,1}\|\hat{\bdelta}_T-\bar{\bdelta}_T\|_1\sqrt{\frac{\log p}{N_{\aux}}}\|\hat{\bb}_T-\bb_T\|_1,
\end{align*}
where the first term comes from the KKT conditions associated with the first step, the second term comes from the sub-Gaussian nature of $\bw^{(0)}_{T,i} $, and the last term comes from the sub-Gaussian nature of $\{\bw^{(k)}_{T,i}\}_{k\in[K]}$.

Notice that
\[
\|\bSig^{(0)}_T\bar{\bSig}_T^{-1}\|_{\infty,1}\leq 1+\|(\bSig^{(0)}_T-\bar{\bSig}_T)\bar{\bSig}_T^{-1}\|_{\infty,1} \leq 1+C_{\Sig}\|\bSig^{(0)}_T\bar{\bSig}_T^{-1}\|_{\infty,1}.
\]
By (\ref{bT-l1}), $\|\hat{\bb}_T-\bb_T\|_1\leq C$ under the sample size condition of Theorem \ref{thm1-tl}.
In event $\mathcal{E}_T$, we have $E_{2,T}\leq C'\lam_0\|\hat{\bdelta}_T-\bar{\bdelta}_T\|_1$. Hence, in $\mathcal{E}_T$, we arrive at the following oracle inequality of $\hat{\bdelta}_T$:
\begin{align*}
\frac{1}{n_0}\|\bW^{(0)}_T(\hat{\bdelta}_T-\bar{\bdelta}_T)\|_2^2 
& \leq \frac{1}{2}\lam_0\|\hat{\bdelta}_T-\bar{\bdelta}_T\|_1+C\sqrt{\frac{\log p}{n_0}}\|\hat{\bdelta}_T-\bar{\bdelta}_T\|_1 \\
& +\lam_0\|\bar{\bdelta}_T\|_1-\lam_0\|\hat{\bdelta}_T\|_1.   
\end{align*}
In $\mathcal{E}_T$, with the sample size condition and the choice of $\lam_0$ in  Lemma \ref{lem1-tl}, standard arguments lead to
\begin{align*}
\|\hat{\bdelta}_T-\bar{\bdelta}_T\|_2^2\vee \frac{1}{n_0}\|\bW^{(0)}_T(\hat{\bdelta}_T-\bar{\bdelta}_T)\|_2^2&\leq Ch\sqrt{\frac{\log p}{n_0}}\\
\|\hat{\bdelta}_T-\bar{\bdelta}_T\|_1&\leq Ch.
\end{align*}
Notice that $P(\mathcal{E}_T)\geq \exp\{-c_1\log p\}$.

To get the convergence rate of $\hat{\btheta}_T=\check{\bb}_T+\check{\bdelta}_T$, we need to analyze the thresholded version of $\hat{\bb}_T$ and $\hat{\btheta}_T$. Let $S_T=supp(\btheta_t)\cup \{j: |\bar{\bdelta}_T|\geq \sqrt{\log p/n_0}\}$. Next, we show that in event $\mathcal{E}_T$,
\begin{equation}
    \begin{aligned}
  &\|\check{\bb}_T-\bb_T\|_2^2\lesssim_{\PP} \frac{s\log p}{N_{\aux}}+h\sqrt{\frac{\log p}{n_0}}\wedge h^2~\text{and}~~\|\check{\bb}_T\|_0\leq C|S_T|\nonumber\\
  & \|\check{\bdelta}_T-\bar{\bdelta}_T\|_2^2\leq \|\hat{\bdelta}_T-\bar{\bdelta}_T\|_2^2+h\sqrt{\log p/n_0}~\text{and}~~  \|\check{\bdelta}_T-\tilde{\bdelta}_T\|_0\leq C|S_T|.\label{thres}
  \end{aligned}
\end{equation}
Notice that $\lambda_{\aux}\asymp C\sqrt{\frac{\log p}{N_{\aux}}}+\frac{h}{s\sqrt{\log p/n_0}+h}\sqrt{\frac{\log p}{n_0}}$ and $|S_T|\leq s+h/\sqrt{\log p/n_0}$. Therefore,
\begin{align*}
\|\check{\bb}_T-\bb_T\|_2^2&\leq \|(\check{\bb}_T-\bb_T)_{S_T}\|_2^2+\|(\check{\bb}_T-\bb_T)_{S_T^c}\|_2^2\\
&\leq \|(\hat{\bb}_T-\bb_T)_{S_T}\|_2^2+\lambda_{\aux}^2|S_T|+\|(\check{\bb}_T)_{S_T^c}\|_2^2+\|(\bb_T)_{S_T^c}\|_2^2\\
&\leq \|(\hat{\bb}_T-\bb_T)_{S_T}\|_2^2+\lambda_{\aux}^2|S_T|+\|(\hat{\bb}_T-\bb_T)_{S_T^c}\|_2^2+2\|(\bb_T)_{S_T^c}\|_2^2\\
&\leq \|\hat{\bb}_T-\bb_T\|_2^2+\lambda_{\aux}^2|S_T|+2\|(\bb_T)_{S_T^c}\|_2^2\\
&\lesssim  s\lam_{\aux}^2+h\lam_{\aux}\wedge h^2+(s+\frac{h}{\sqrt{\log p/n_0}})\lam_{\aux}^2+h\sqrt{\frac{\log p}{n_0}}\wedge h^2.
\end{align*}
Notice that $\lam_{\aux}=o(\sqrt{\frac{\log p}{n_0}})$ as long as $h\ll s\sqrt{\frac{\log p}{n_0}}$. Hence,
\begin{align*}
\|\check{\bb}_T-\bb_T\|_2^2&\lesssim  s\lam_{\aux}^2+h\sqrt{\frac{\log p}{n_0}}\wedge h^2\\
&\lesssim \frac{s\log p}{N_{\aux}}+h\sqrt{\frac{\log p}{n_0}}.
\end{align*}
Moreover,
\begin{align*}
   \|\check{\bb}_T\|_0&\leq |S_T|+\sum_{j\notin S_T} \mathbbm{1}(|(\hat{\bb}_T)_j|\geq \lambda_{\aux})\\
   &\leq |S_T|+\sum_{j\notin S_T,(\bb_T)_j= 0} \mathbbm{1}(|(\hat{\bb}_T-\bb_T)_j|\geq \lambda_{\aux})+|\{j\notin S_T,(\bb_T)_j\neq 0\}|\\
   &\lesssim C|S_T|.
\end{align*}
For the thresholded version $\check{\bdelta}_T$, it is easy to show that $\|\tilde{\bdelta}_T\|_0\leq h/\sqrt{\log p/n_0}$. The proof follows similarly.
Finally, notice that
\begin{align*}
\|\hat{\btheta}_t-\btheta_t+(\bar{\bdelta}_T-\tilde{\bdelta}_T)\|_0 
& \leq \|\check{\bb}_T-\bb_T+\check{\bdelta}_T-\tilde{\bdelta}_T\|_0 \\
& \leq C|S_T|+\|\check{\bdelta}_T-\tilde{\bdelta}_T\|_0 \\
& \leq C'|S_T|.
\end{align*}
The proof is complete.
\end{proof}

\begin{proof}[Proof of Theorem \ref{thm1-tl}]
We are left to prove the convergence rate of $\hat{\btheta}_t-\btheta_t$ for $t<T$. 

For $\bb_t$ defined in \eqref{eq-bt}, the following oracle inequality holds
\begin{align*}
&\frac{1}{N_{\aux}}\sum_{k=1}^K\|\bW_t^{(k)}(\hat{\bb}_t-\bb_t)\|_2^2 \\ 
& \leq\frac{1}{N_{\aux}}|\sum_{k=1}^K\langle \bW_t^{(k)}(\hat{\bb}_t-\bb), \hat{\by}_t^{(k)}-\bW_t^{(k)}\bb_t)\rangle| +\lambda_{\aux}(\|\bb_{t}\|_1-\|\hat{\bb}_t\|_1) \\
&\leq\frac{1}{N_{\aux}} |\sum_{k=1}^K\langle \bW_t^{(k)}(\hat{\bb}_t-\bb), \br_t^{(k)}+\max_{\ba}Q_{t+1}(\bW_{t+1}^{(k)},\ba;\btheta_{t+1})-\bW_t^{(k)}\bb_t\rangle| \\
& +\lambda_{\aux}(\|\bb_{t}\|_1-\|\hat{\bb}_t\|_1)\\
& +\underbrace{ \frac{1}{N_{\aux}}\sum_{k=1}^K\|\max_{\ba}Q_{t+1}(\bW_{t+1}^{(k)},\ba;\hat{\btheta}_{t+1})-\max_{\ba}Q_{t+1}(\bW_{t+1}^{(k)},\ba;\btheta_{t+1})\|_2^2}_{E_{1,t}},
\end{align*}
where $\br_t^{(k)}+\max_{\ba}Q_{t+1}(\bW_{t+1}^{(k)},\ba;\btheta_{t+1})=\by_t^{(tl-k)}$ by definition and we use the definition of $\hat{\by}_t^{(k)}$ and  $\by_t^{(k)}$ in the last step.
Using the second statement in $\mathcal{E}_t$, we have
\begin{align*}
&\frac{1}{N_{\aux}}\sum_{k=1}^K\|\bW_t^{(k)}(\hat{\bb}_t-\bb_t)\|_2^2\leq  \frac{1}{N_{\aux}}|\sum_{k=1}^K\langle \bW_t^{(k)}(\hat{\bb}_t-\bb), \hat{\by}_t^{(k)}-\bW_t^{(k)}\bb_t)\rangle| \\
&+\lambda_{\aux}(\|\bb_{t}\|_1-\|\hat{\bb}_t\|_1)\\
&\leq \frac{\lambda_{\aux}}{2}\|\hat{\bb}_t-\bb_t\|_1+ \lambda_{\aux}\|\bb_{t}\|_1-\lambda_{\aux}\|\hat{\bb}_t\|_1 +E_{1,t}\\
&\leq \frac{3\lambda_{\aux}}{2}\|(\hat{\bb}_t-\bb_t)_{\text{Supp}(\bb_t)}\|_1-\frac{\lambda_{\aux}}{2}\|(\hat{\bb}_t-\bb_t)_{\text{Supp}^c(\bb_t)}\|_1+2\lam_{\aux}\|(\bb_t)_{\text{Supp}^c(\bb_t)}\|_1 \\
& + E_{1,t}.
\end{align*}
Similarly to the proof of Theorem 1 in \cite{li2022transfer-jrssb}, we can show that under the conditions of Theorem \ref{thm1-tl}, if $E_{1,t}=o(1)$, then
\[
 \frac{1}{N_{\aux}}\sum_{k=1}^K\|\bW_t^{(k)}(\hat{\bb}_t-\bb_t)\|_2^2\vee \|\hat{\bb}_t-\bb_t\|_2^2\lesssim s\lam_{\aux}^2+h\lam_{\aux}+E_{1,t}.
\]

For $E_{1,t}$, we have
\begin{align*}
E_{1,t}&\leq \frac{1}{N_{\aux}}\sum_{k=1}^K \|\bX_{t+1}^{(k)}(\hat{\btheta}_{t+1,1:p}-\btheta_{t+1,1:p})\|_2^2 \\
& +\frac{1}{N_{\aux}}\sum_{k=1}^K \|\bX_{t+1}^{(k)}(\hat{\btheta}_{t+1,(p+1):2p}-\btheta_{t+1,(p+1):2p})\|_2^2\\
&\leq \frac{2}{N_{\aux}}\sum_{k=1}^K\|\bW_{t+1}^{(k)}(\hat{\btheta}_{t+1}-\btheta_{t+1})\|_2^2\\
&\leq\frac{4}{N_{\aux}}\sum_{k=1}^K\|\bW_{t+1}^{(k)}(\hat{\btheta}_{t+1}+(\bar{\bdelta}_{t+1}-\tilde{\bdelta}_{t+1})-\btheta_{t+1})\|_2^2 \\
&+\frac{4}{N_{\aux}}\sum_{k=1}^K\|\bW_{t+1}^{(k)}(\bar{\bdelta}_{t+1}-\tilde{\bdelta}_{t+1})\|_2^2.
\end{align*}

In the sequel, we provide the proof for $t=T-1$. The results for previous stages hold by induction.
We have shown in Lemma \ref{lem1-tl} that $\|\hat{\btheta}_T+(\bar{\bdelta}_T-\tilde{\bdelta}_T)-\btheta_T\|_0\leq C|S_T|$. Hence, we can use the upper restricted eigenvalue condition in $\mathcal{E}_t$ to arrive at
\[
  E_{1,t}\lesssim\|\hat{\btheta}_{T}-\btheta_T\|_2^2+\|\bar{\bdelta}_T-\tilde{\bdelta}_T\|_2^2,
\]
with probability at least $1-\exp(-c_1\log p)$ given that $|S_T|\log p=o(N_{\aux})$. 

To summarize, under the current sample size condition, with probability at least $1-\exp(-c_1\log p)$,
\begin{align}
&\frac{1}{N_{\aux}}\sum_{k=1}^K\|W_t^{(k)}(\hat{\bb}_t-\bb_t)\|_2^2\vee \|\hat{\bb}_t-\bb_t\|_2^2\lesssim \frac{s\log p}{N_{\aux}}+h\sqrt{\frac{\log p}{N_{\aux}}}+\|\hat{\btheta}_{t+1}-\btheta_{t+1}\|_2^2\\
&\|\hat{\bb}_t-\bb_t\|_1\lesssim s\lam_{\aux}+h+\frac{\|\hat{\btheta}_{t+1}-\btheta_{t+1}\|_2^2}{\lam_{\aux}}\lesssim s\sqrt{\frac{\log p}{N_{\aux}}}+h+\sqrt{hs}(\frac{log p}{n_0})^{1/4}.\label{bt-l1}
\end{align}

For $\hat{\bdelta}_t-\bar{\bdelta}_t$, we can similarly derive that
\begin{align*}
\frac{1}{n_0}\|\bW^{(0)}_t(\hat{\bdelta}_t-\bar{\bdelta}_t)\|_2^2&\leq \frac{1}{n_0}\langle \bW^{(0)}_t(\hat{\bdelta}_t-\bar{\bdelta}_t),r_t^{(0)}-\bW_t^{(0)}(\hat{\bb}_t+\bar{\bdelta}_t)\rangle \\
& +\lam_0\|\bar{\bdelta}_t\|_1-\lam_0\|\hat{\bdelta}_t\|_1\\
&\leq |\frac{1}{n_0}\langle \bW^{(0)}_t(\hat{\bdelta}_t-\bar{\bdelta}_t),r_t^{(0)}-\bW_t^{(0)}\btheta_t\rangle| \\
& +\underbrace{2(\hat{\bdelta}_t-\bar{\bdelta}_t)^\top\widehat{\bSig}^{(0)}_t(\hat{\bb}_t-\bb_t)}_{E_{2,t}}+\lam_0\|\bar{\bdelta}_t\|_1-\lam_0\|\hat{\bdelta}_t\|_1
\end{align*}
For $E_{2,t}$, analogous to the proof of Lemma \ref{lem1-tl}, we have in $\mathcal{E}_t$,
\begin{align*}
E_{2,t}\leq  \|\hat{\bdelta}_t-\bar{\bdelta}_t\|_1C\sqrt{\frac{\log p}{n_0}} (1+\|\hat{\bb}_t-\bb_t\|_1).
\end{align*}
In view of (\ref{bt-l1}),  $\|\hat{\bb}_t-\bb_t\|_1\leq C$ under the sample size conditions of Theorem \ref{thm1-tl}.
 Therefore, in the event $\mathcal{E}_t$, we have
\begin{align*}
  \|\hat{\bdelta}_t-\bar{\bdelta}_t\|_2^2\leq Ch\sqrt{\frac{\log p}{n_0}}.
\end{align*}

Let $S_t=supp(\btheta_t)\cup \{j: |\bar{\bdelta}_t|\geq \sqrt{\log p/n_0}\}$. Next, we can similarly show that
\begin{equation}
\label{thres-t}
  \|\check{\bb}_t-\bb_t\|_2^2\lesssim_P \frac{s\log p}{N_{\aux}}+h\sqrt{\frac{\log p}{n_0}}\wedge h^2~\text{and}~\|\check{\bb}_t\|_0=O_P(|S_t|).
\end{equation}

For the thresholded version $\check{\bdelta}_t$,
we similarly have
\[
    \|\check{\bdelta}_t-\bar{\bdelta}_t\|_2^2\leq \|\hat{\bdelta}_t-\bar{\bdelta}_t\|_2^2+h\sqrt{\log p/n_0}
\]
and
\[
  \|\check{\bdelta}_t-\tilde{\bdelta}_t\|_0\leq C|S_t|.
\]
where $\tilde{\bdelta}_t$ is defined as a thresholded version of $\bdelta_t$ at threshold level $\lam_0$.

\end{proof}

\begin{proof}[Proof of Theorem \ref{thm-ol}]
Using the relationship between the $Q$-function and the value function, i.e., (2.1) and (2.2) in \cite{hao2021online}, we have for the exploitation phase
\begin{align*}
\sum_{i=1}^N\gamma^t\sum_{t=1}^Tr_t^{(0)}(\bx^{(0)}_{t,i},a^*_{t,i})&=\max_{a\in\{-1,1\}}Q_1(\bx^{(0)}_{1,i},a)=\bx^{(0)}_{1,i}\bbeta_1+\gamma|\bx^{(0)}_{1,i}\bpsi_1|\\
\sum_{i=n_{\ex}+1}^N\gamma^t\sum_{t=1}^Tr_t^{(0)}(\bx^{(0)}_{t,i},\hat{a}^{(0)}_{t,i}) &=Q_1(\bx_{1,i}^{(0)},\hat{a}_{1,i}^{(0)})=\bx^{(0)}_{1,i}\hat{\bbeta}_1+\gamma|\bx^{(0)}_{1,i}\hat{\bpsi}_1|
\end{align*}

\begin{align*}
&\sum_{i=1}^N\gamma^t \sum_{t=1}^T(r_t^{(0)}(\bx^{(0)}_{i,t},a^*_{i,t})-r_t^{(0)}(\bx^{(0)}_{i,t},\hat{a}_{i,t}))\\
&\leq n_{\ex}\frac{\gamma-\gamma^T}{1-\gamma}+\sum_{i=n_{\ex}+1}^N\bx_{1,i}(\bbeta_1-\hat{\bbeta}_1)+\gamma(|\bx^{(0)}_{1,i}\bpsi_1|-|\bx^{(0)}_{1,i}\hat{\bpsi}_1|)\\
&\leq \frac{n_{\ex}\gamma}{1-\gamma}+\sum_{i=n_{\ex}+1}^I |\bx_{1,i}^{(0)}(\hat{\btheta}_1-\btheta_1)|.
\end{align*}
As $\hat{\btheta}_1-\btheta_1$ is independent of $\bx^{(0)}_{1,i}$ for $i> n_{\ex}$, conditioning on $\hat{\btheta}_1-\btheta_1$, $|\bx^{(0)}_{1,i}(\hat{\bpsi}_1-\bpsi_1)|$ is \textit{i.i.d.} sub-Gaussian with sub-Gaussian norm $C\|\hat{\btheta}_1-\btheta_1\|_2$. Hence,
\begin{align*}
\sum_{i=n_{\ex}+1}^I |\bx_{1,i}^{(0)}(\hat{\btheta}_1-\btheta_1)|\leq (N-n_{\ex}) \|\hat{\btheta}_1-\btheta_1\|_2+\sqrt{N-n_{\ex}} \|\hat{\btheta}_1-\btheta_1\|_2,
\end{align*}
with probability at least $1-\exp\{-(N-n_{\ex})\}$.

\end{proof}

\underline{\textbf{Proof of (\ref{reg-tl-opt})}}.
By (\ref{reg-tl}),
\begin{align*}
{\rm Regret}_{NT}
\lesssim
\frac{n_{\ex}\gamma }{1-\gamma}
+
(N-n_{\ex}) \paran{\sqrt{\frac{s\log p}{N_{\aux}}}+ h^{1/2} \paran{\frac{\log p}{n_{\ex}}}^{1/4}}:=b_N.
\end{align*}
We have
\begin{align*}
b_N\lesssim n_{\ex}+N\paran{\sqrt{\frac{s\log p}{N_{\aux}}}+ h^{1/2} \paran{\frac{\log p}{n_{\ex}}}^{1/4}}.
\end{align*}
Then the optimal $n_{\ex}$ satisfies
\[
    n_{\ex}\asymp Nh^{1/2}(\frac{\log p}{n_{\ex}})^{1/4}\implies n_{\ex}\asymp (Nh^{1/2}(\log p)^{1/4})^{4/5}.
\]

To summarize, we see that it suffices to take 
$$n_{\ex} \asymp   \max\{N^{4/5}h^{2/5}(\log p)^{1/5},s^2h^2\log p\}$$ 
and the corresponding regret is of order
\[
  \max\{N^{4/5}h^{2/5}(\log p)^{1/5},s^2h^2\log p\}+N\sqrt{\frac{s\log p}{N_{\aux}}}.
\]

\section{Further results on numerical experiments}

The true coefficients for the Q-functions in (\ref{eqn:true-q}) are $\theta_{2j} = \kappa_j$, $1\le j\le 7$ and
\begin{equation}  \label{eqn:true-q-theta}
    \begin{aligned}
       \theta_{11} & = \kappa_1 + q_1 \abs{f_1} + q_2 \abs{f_2} + (0.5-q_1) \abs{f_3} + (0.5-q_2)\abs{f_4},  \\
       \theta_{12} & = \kappa_2 + q_1'\abs{f_1} + q_2' \abs{f_2} - q_1' \abs{f_3} - q_2' \abs{f_4}, \\
       \theta_{13} & = \kappa_3 + q_1 \abs{f_1} - q_2 \abs{f_2} + (0.5-q_1) \abs{f_3} - (0.5-q_2)\abs{f_4}, \\
       \theta_{14} & = \kappa_4 + q_1'\abs{f_1} - q_2' \abs{f_2} - q_1' \abs{f_3} + q_2' \abs{f_4},
    \end{aligned}
\end{equation}
where 
\begin{align*}
q_1 & = 0.25\paran{ {\rm expit}\paran{b_1 + b_2} + {\rm expit}\paran{-b_1 + b_2}} \\   
q_2 & = 0.25\paran{ {\rm expit}\paran{b_1 - b_2} + {\rm expit}\paran{-b_1 - b_2} } \\
q_1'& = 0.25\paran{ {\rm expit}\paran{b_1 + b_2} - {\rm expit}\paran{-b_1 + b_2} } \\
q_2'& = 0.25\paran{ {\rm expit}\paran{b_1 - b_2} - {\rm expit}\paran{-b_1 - b_2} } \\
f_1 & = \kappa_5 + \kappa_6 + \kappa_7 \\
f_2 & = \kappa_5 + \kappa_6 - \kappa_7 \\
f_3 & = \kappa_5 - \kappa_6 + \kappa_7 \\
f_4 & = \kappa_5 - \kappa_6 - \kappa_7
\end{align*}


\section{Variables in MIMIC-III}

The covariates used in real data applications are 'gender', 'age', 'elixhauser', 're\_admission', 'Weight\_kg','GCS', 'HR', 'SysBP', 'MeanBP', 'DiaBP', 'RR', 'SpO2', 'Temp\_C', 'FiO2\_1', 'Potassium',
'Sodium', 'Chloride', 'Glucose', 'BUN', 'Creatinine', 'Magnesium', 'Calcium',
'Ionised\_Ca', 'CO2\_mEqL', 'SGOT', \\
'SGPT', 'Total\_bili', 'Albumin', 'Hb', 'WBC\_count',
'Platelets\_count', 'PTT', 'PT', 'INR', 'Arterial\_pH', 'paO2', 'paCO2', 'Arterial\_BE',
'Arterial\_lactate', 'HCO3', 'Shock\_Index', 'PaO2\_FiO2', 'SOFA', 'SIRS'.
The meaning of those variables are detailed in Table \ref{tab:variables}.

\begin{table}[tb!]
    \centering
    \caption{Variables in MIMIC-III}
    \label{tab:variables}
    \resizebox{\textwidth}{!}{%
        \begin{tabular}{|l|l|l|l|}
            \hline
            \multicolumn{1}{|c|}{} & \multicolumn{1}{c|}{Item} & \multicolumn{1}{c|}{Header} & \multicolumn{1}{c|}{Type} \\ \hline
            Demographics & Age & age & Integer \\
            & Gender & gender & Binary \\
            & Weight & Weight\_kg & Continuous \\
            & Readmission to ICU & re\_admission & Binary \\
            & Elixhauser score (premorbid status) & elixhauser & Continuous \\ \hline
            Vital signs & Modified SOFA & SOFA & Continuous \\
            & SIRS & SIRS & Continuous \\
            & Glasgow coma scale & GCS & Continuous \\
            & Heart rate & HR & Continuous \\
            & Systolic blood pressure & SysBP & Continuous \\
            & Mean blood pressure & MeanBP & Continuous \\
            & Diastolic blood pressure & DiaBP & Continuous \\
            & Shock index & Shock\_Index & Continuous \\
            & Respiratory rate & RR & Continuous \\
            & SpO2 & SpO2 & Continuous \\
            & Temperature & Temp\_C & Continuous \\ \hline
            Lab values & Potassium & Potassium & Continuous \\
            & Sodium & Sodium & Continuous \\
            & Chloride & Chloride & Continuous \\
            & Glucose & Glucose & Continuous \\
            & BUN & BUN & Continuous \\
            & Creatinine & Creatinine & Continuous \\
            & Magnesium & Magnesium & Continuous \\
            & Calcium & Calcium & Continuous \\
            & Ionized calcium & Ionised\_Ca & Continuous \\
            & Carbon dioxide & CO2\_mEqL & Continuous \\
            & SGOT & SGOT & Continuous \\
            & SGPT & SGPT & Continuous \\
            & Total bilirubin & Total\_bili & Continuous \\
            & Albumin & Albumin & Continuous \\
            & Hemoglobin & Hb & Continuous \\
            & White blood cells count & WBC\_count & Continuous \\
            & Platelets count & Platelets\_count & Continuous \\
            & PPT & PTT & Continuous \\
            & PT & PT & Continuous \\
            & INR & INR & Continuous \\
            & pH & Arterial\_pH & Continuous \\
            & PaO2 & paO2 & Continuous \\
            & PaCO2 & paCO2 & Continuous \\
            & Base excess & Arterial\_BE & Continuous \\
            & Bicarbonate & HCO3 & Continuous \\
            & Lactate & Arterial\_lactate & Continuous \\
            & PaO2/FiO2 ratio & PaO2\_FiO2 & Continuous \\ \hline
            Ventilation parameters & Mechanical ventilation & mechvent & Binary \\
            & FiO2 & FiO2\_1 & Continuous \\ \hline
            Medications and fluid balance & Current IV fluid intake over 4h & median\_dose\_vaso & Continuous \\
            & Maximum dose of vasopressor over 4h & max\_dose\_vaso & Continuous \\
            & Urine output over 4h & output\_4hourly & Continuous \\
            & Cumulated fluid balance since admission (includes readmission data when available) & cumulated\_balance & Continuous \\ \hline
            Outcome & Hospital mortality & died\_in\_hosp & Binary \\
            & 90-day mortality & died\_within\_48h\_of\_out\_time,mortality\_90d & Binary \\ \hline
        \end{tabular}%
    }
\end{table}

\begin{table}[tb!]
    \caption{\label{tab:horizons}Horizon lengths and occurrences.}
    \centering
        \begin{tabular}{ccccccccccc}
            \hline
            Length & 1 & 2 & 3 & 4 & 5 & 6 & 7 & 8 & 9 & 10  \\
            \hline
            \# Traces &  271 &  170 &  177 &  184 &  260 &  485 &  879 &  1142 &  1371 &  1331  \\
            \hline
\hline
            Length & 11 & 12 & 13 & 14 & 15 & 16 & 17 & 18 & 19 & 20 \\
            \hline
             \# Traces &  1458 &  1573 &  1500 &  1694 &  1429 &  1041 &  789 &  659 &  524 &  4006 \\
            \hline
        \end{tabular}
\end{table}

\section{More tables and plots}

Table \ref{tab:least} presents the settings with least number of samples (top 10) in our MIMIC-III data.
The settings are defined by different combinations of ``Gender'' $g$, ``Inverse step'' $T-t$ , and action ``IV fluid'' $\times$ ``Vasopressor''.

\begin{table}[tb!]
   \caption{\label{tab:least}Settings with least number of samples (top 10) in our MIMIC-III data.
   The settings are defined by different combinations of ``Gender'' $g$, ``Inverse step'' $T-t$ , and action ``IV fluid'' $\times$ ``Vasopressor''. } 
    \centering
    \begin{tabular}{ccccc}
        \hline
        Gender & Inverse step & IV fluid & Vaso &  \# Samples  \\
        \hline
        0 & 19 & 1 & 1 &  18 \\
        1 & 19 & 0 & 1 &  19 \\
        0 & 19 & 0 & 1 &  20 \\
        1 & 18 & 0 & 1 &  20 \\
        1 & 19 & 1 & 2 &  26 \\
        1 & 19 &  1  & 1 &  26 \\
        1 & 17 & 0 & 1 &  26 \\
        1 & 17  & 1 & 1 &  27 \\
        1 & 16 & 1 & 2 &  28 \\
        1 & 18 & 1 & 1 &  28 \\
        \hline
    \end{tabular}
\end{table}

\section{Description of MIMIC-III Dataset}  \label{appen:mimic-iii}

\subsection{Intensive Care Unit Data}

The data we use is the Medical Information Mart for Intensive Care version III (MIMIC-III) Database \citep{johnson2016mimic}, which is a freely available source of de-identified critical care data from 53,423 adult admissions and 7,870 neonates from 2001 -- 2012 in six ICUs at a Boston teaching hospital.
The database contain high-resolution patient data, including demographics, time-stamped measurements from bedside monitoring of vital signs, laboratory tests, illness severity scores, medications and procedures, fluid intakes and outputs, clinician notes and diagnostic coding.

We extract a cohort of sepsis patients, following the same data processing procedure as in \cite{komorowski2018artificial}.
Specifically, the adult patients included in the analysis satisfy the international consensus sepsis-3 criterion.
The data includes 17,083 unique ICU admissions from five separate ICUs in one tertiary teaching hospital.
Patient demographics and clinical characteristics are shown in Table 1 and Supplementary Table 1 of \cite{komorowski2018artificial}.

Each patient in the cohort is characterized by a set of 47 variables, including demographics, Elixhauser premorbid status, vital signs, and laboratory values.
Demographic information includes age, gender, weight. Vital signs include heart rate, systolic/diastolic blood pressure, respiratory rate et al.
Laboratory values include glucose, total bilirubin, (partial) thromboplastin time et al.
Patients' data were coded as multidimensional discrete time series with 4-hour time steps.
The actions of interests are the total volume of intravenous (IV) fluids and maximum dose of vasopressors administrated over each 4-hour period.

All features were checked for outliers and errors using a frequency histogram method and uni-variate statistical approaches (Tukey's method).
Errors and missing values are corrected when possible.
For example, conversion of temperature from Fahrenheit to Celsius degrees and capping variables to clinically plausible values.

In the final processed data set, we have {17621} unique ICU admissions, corresponding to unique trajectories fed into our algorithms.

\subsubsection{Irregular observational time series data}

For each ICU admission, we code patient's data as multivariate discrete time series with a four hour time step.
Each trajectory covers from up to 24h preceding until 48h following the estimated onset of sepsis, in order to capture the early phase of its management, including initial resuscitation.
The medical treatments of interest are the total volume of intravenous fluids and maximum dose of vasopressors administered over each four hour period.
We use a time-limited parameter specific sample-and-hold approach to address the problem of missing or irregularly sampled data.
The remaining missing data were interpolated in MIMIC-III using multivariate nearest-neighbor imputation.
After processing, we have in total {278598} sampled data points for the entire sepsis cohort.

\subsubsection{State and action space characterization}

The state $\bX_{i,t}$ is a 47-dimensional feature vector including fixed demographic information (age, weight, gender, admit type, ethnicity et al), vitals signs (heart rate, systolic/diastolic blood pressure, respiratory rate et al), and laboratory values (glucose, Creatinine, total bilirubin, partial thromboplastin time, $paO_2$, $paCO_2$ et al.).

For action space, we discretize two variables into three actions respectively according to Table in \cite{komorowski2018artificial}.
The combination of the two drugs makes $3 \times 3 = 9$ possible actions in total.
The action $A_t$ is a two-dimensional vector, of which the first entry $a_t[0]$ specifies the dosages of IV fluids and the second $a_t[1]$ indicates the dosages of IV fluids and vasopressors, to be administrated over the next 4h interval.

\subsection{Reward design}

The reward signal is important and needs to be crafted carefully in real applications.  \cite{komorowski2018artificial} uses hospital mortality or 90-day mortality as the sole defining factor for the penalty and reward.
Specifically, when a patient survived, a positive reward was released at the end of each patient's trajectory (a reward of {+ 100}); while a negative reward (a penalty of {-100}) was issued if the patient died.
However, this reward design is sparse and provides little information at each step. Also, mortality may be correlated with the health statues of a patient. So it is reasonable to associate reward to the health measurement of a patient after an action is taken.

In this application, we build our reward signal based on physiological stability.
Specifically, in our design, physiological stability is measured by vitals and laboratory values $v_t$ with desired ranges $[ v_{\min}, v_{\max} ]$.
Important variables related to sepsis include heart rate (HR), systolic blood pressure (SysBP), mean blood pressure (MeanBP), diastolic blood pressure (DiaBP), respiratory rate (RR), peripheral capillary oxygen saturation (SpO2), arterial lactate, creatinine, total bilirubin, glucose, white blood cell count, platelets count, (partial) thromboplastin time (PTT), and International Normalized Ratio (INR).
We encode a penalty for exceeding desired ranges at each time step by a truncated Sigmoid function, as well as a penalty for sharp changes in consecutive measurements.


Here, values $v_t$ are the measurements of those vitals $v$ believed to be indicative of physiological stability at time $t$, with desired ranges $[v_{min}, v_{max}]$. The penalty for exceeding these ranges at each time step is given by a truncated sigmoid function. The system also receives negative feedback when consecutive measurements see a sharp change.

\begin{remark}
    There are definitely improvements in shaping the reward space. For example, in medical situation, the definition of the normal range of a variable sometime depends demographic characterization. Also, sharp changes in a favorable direction should be rewarded.
\end{remark}

\end{appendix}

\end{document}